# Optimal Time Complexity Algorithms for Computing General RWKs on Sparse Graphs

**Krzysztof Choromanski**[1 2*], **Isaac Reid**[3 4*], **Arijit Sehanobish**[5], **Avinava Dubey**[4]
[1]Google DeepMind, [2]Columbia University, [3]University of Cambridge, [4]Google Research, [5]Independent researcher

## Abstract

We present the first linear time complexity randomized algorithms for unbiased approximation of the celebrated family of general *random walk kernels* (RWKs) for sparse graphs. This includes both labelled and unlabelled instances. The previous fastest methods for general RWKs were of cubic time complexity and not applicable to labelled graphs. Our method samples dependent random walks to compute novel graph embeddings in $\mathbb{R}^d$ whose dot product is equal to the true RWK in expectation. It does so without instantiating the direct product graph in memory, meaning we can scale to massive datasets that cannot be stored on a single machine. We derive exponential concentration bounds to prove that our estimator is sharp, and show that the ability to approximate general RWKs (rather than just special cases) unlocks efficient implicit graph kernel learning. Our method is up to **27**× faster than its counterparts for efficient computation on large graphs and scales to graphs **128**× bigger than largest examples amenable to brute-force computation.

## 1 INTRODUCTION

Consider the family of *graph kernels* $\mathrm{K} : \mathcal{G} \times \mathcal{G} \to \mathbb{R}$, positive definite, symmetric functions which assign real similarity scores to pairs of input graphs. Whilst their counterparts in Euclidean space have enjoyed widespread adoption across machine learning – including for Gaussian processes (Williams and Rasmussen, 2006), clustering (Cristianini et al., 2001; Liu et al., 2020; Liang et al., 2024), 3d-reconstruction (Williams et al., 2022; Huang et al., 2023) and linear attention transformers (Choromanski et al., 2020; Likhosherstov et al., 2022) – graph kernel methods have remained underutilized. A major reason for this is their prohibitive computational cost. Letting $N$ denote the number of graph vertices, exact methods are typically $\mathcal{O}(N^6)$, or at best $\mathcal{O}(N^3)$ with constraints and approximations.

**Random walk kernels.** Consider an unweighted, undirected graph $\mathrm{G}(\mathrm{V}, \mathrm{E})$ where $\mathrm{V} \coloneqq \{v_1, ..., v_N\}$ is the set of vertices and $\mathrm{E}$ is the set of edges, with $(v_i, v_j) \in \mathrm{E}$ if and only if there exists an edge between $v_i$ and $v_j$ in $\mathrm{G}$. Given a pair of graphs $\mathrm{G}_1, \mathrm{G}_2$, a popular, general parameterization of kernels $\mathrm{K}$ is the class of general *random walk kernels* (RWKs),

$$\mathrm{K}_{\mathrm{RWK}}(\mathrm{G}_1, \mathrm{G}_2) \stackrel{\mathrm{def}}{=} \mathbf{v}^\top \left[ \sum_{i=0}^{\infty} \mu_i \mathbf{A}^i_{\mathrm{G}_1 \times \mathrm{G}_2} \right] \mathbf{w}. \quad (1)$$

Here, $\mathrm{G}_1 \times \mathrm{G}_2$ denotes the *direct product* of $\mathrm{G}_1$ and $\mathrm{G}_2$. Given a vertex labelling function $\mathcal{L} : \mathrm{V}(\mathrm{G}_1) \cup \mathrm{V}(\mathrm{G}_2) \to \mathrm{L}$ (for a discrete set of labels $\mathrm{L}$), this is obtained by taking $\mathrm{V}(\mathrm{G}_1 \times \mathrm{G}_2) = \{(v_1, v_2) : v_1 \in \mathrm{V}(\mathrm{G}_1), v_2 \in \mathrm{V}(\mathrm{G}_2), \mathcal{L}(v_1) = \mathcal{L}(v_2)\}$, and $\mathrm{E}(\mathrm{G}_1, \mathrm{G}_2)= \{((v_1, v_2), (v_1', v_2')) : (v_1, v_1') \in \mathrm{E}(\mathrm{G}_1), (v_2, v_2') \in \mathrm{E}(\mathrm{G}_2), (v_1, v_2) \in \mathrm{V}(\mathrm{G}_1 \times \mathrm{G}_2), (v_1', v_2') \in \mathrm{V}(\mathrm{G}_1 \times \mathrm{G}_2)\}$. Here $\mathbf{A}_{\mathrm{G}_1 \times \mathrm{G}_2} \in \mathbb{R}^{N_1 N_2 \times N_1 N_2}$ is the corresponding adjacency matrix, and $(\mu_i)_{i=0}^{\infty}$ is a sequence of coefficients chosen to decay fast enough to ensure convergence. Vectors $\mathbf{v}, \mathbf{w} \in \mathbb{R}^{N_1 N_2}$ encode joint distributions on the vertices of $\mathrm{G}_1 \times \mathrm{G}_2$. They are often factorized as products of independent distributions on $\mathrm{G}_1$ and $\mathrm{G}_2$, so that $\mathbf{v} = \mathrm{flat}(\mathbf{v}_1 \otimes \mathbf{v}_2)$ where flat is the 'vectorizing' operation, $\otimes$ is the outer product, and $\mathbf{v}_1 \in \mathbb{R}^{N_1}$, $\mathbf{v}_2 \in \mathbb{R}^{N_2}$. Intuitively, RWKs measure graphs' similarity by taking a weighted sum over walks of different lengths on $\mathrm{G}_1 \times \mathrm{G}_2$, thereby counting the number of shared walks present in both $\mathrm{G}_1$ and $\mathrm{G}_2$. Vishwanathan et al. (2010) noted that, with special choices for $\mathbf{v}$, $\mathbf{w}$ and $(\mu_i)_{i=1}^{\infty}$, Eq. 1 includes *marginalized graph kernels* (Scholkopf et al., 2005), *geometric random walk kernels* (taking $\mu_i = \lambda^i$), and *exponential kernels* (taking $\mu_i = \frac{\lambda^i}{i!}$ and $\mathbf{v}, \mathbf{w}$ uniform) (Gärtner et al., 2003). Therefore, general RWKs include several popular graph kernels.

Proceedings of the 28th International Conference on Artificial Intelligence and Statistics (AISTATS) 2025, Mai Khao, Thailand. PMLR: Volume 258.

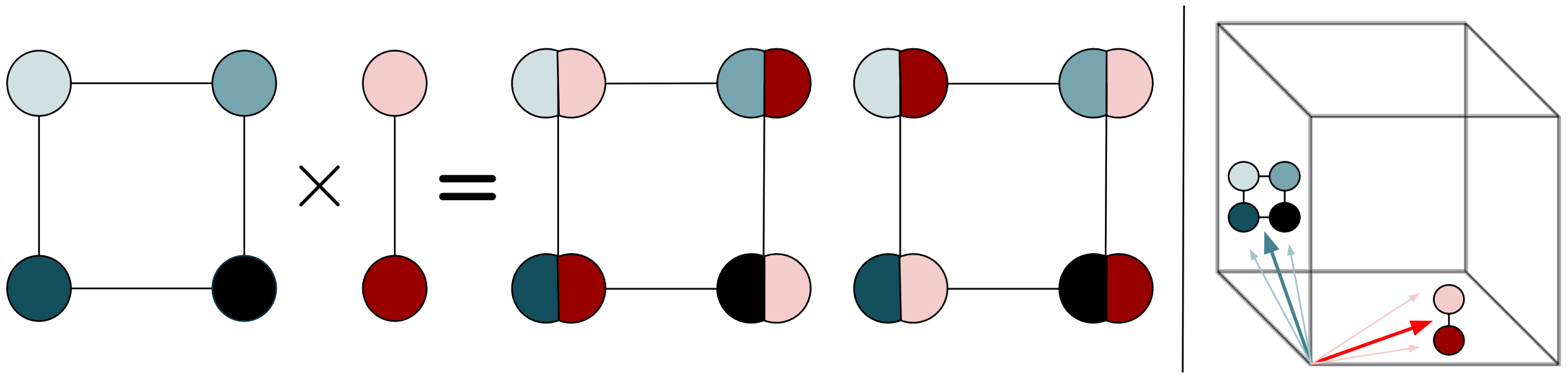

Figure 1: **Left:** an example of the direct product of two graphs, a core concept of RWKs. **Right:** a schematic view of our approach. The graphs are embedded in $d_{\mathrm{G}}$-dimensional Euclidean space (here, $d_{\mathrm{G}} = 3$), such that the dot product of embeddings equals RWKs in expectation. The embeddings are randomized; shadow arrows represent different possible realizations. Computing the embeddings means the direct product graph need not be instantiated in memory.

**Time complexity.** Letting $N_1 = N_2 = N$, brute force computation of RWK kernel evaluations via Eq. 1 incurs time complexity $O(N^6)$. Vishwanathan et al. (2010) proposed alternatives that improve this to $\mathcal{O}(N^3)$ for general RWKs, though only for unlabelled graphs (see Sec. 2). However, for sparse graphs and factorized vectors $\mathbf{v} = \text{flat}(\mathbf{v}_1 \otimes \mathbf{v}_2)$ and $\mathbf{w} = \text{flat}(\mathbf{w}_1 \otimes \mathbf{w}_2)$, the *inputs* to the RWK are only of size $\mathcal{O}(N)$. This provides a lower bound on the time complexity to compute the RWK, and invites the following question:

*Is it possible to compute unbiased estimates of general RWKs for sparse, unweighted, undirected, labelled graphs in time complexity $O(N)$, for arbitrary $(\mu_i)_{i=0}^{\infty}$?*

In this paper, we provide a comprehensive affirmative answer, presenting *graph voyagers* (GVoys): the **first $\mathcal{O}(N)$ algorithm for the unbiased approximation of general RWKs, for labelled and unlabelled graphs**. To our knowledge, GVoys are **first sub-cubic** algorithm to achieve this. Our method generalizes the recently-introduced class of *graph random features* (GRFs; Choromanski, 2023; Reid et al., 2023, 2024b,a), which approximate kernels defined on the *vertices* of graphs. We sample random walks on $\mathrm{G}_1$ and $\mathrm{G}_2$ with shared random variables, thereby emulating walks on $\mathrm{G}_1 \times \mathrm{G}_2$ but without ever instantiating the product graph in memory. The walks are used to generate a set of low-dimensional feature representations of $\{\mathrm{G}_i\}_{i=1}^{N_G}$ (where $N_G$ is the number of graphs in the dataset), such that the Euclidean dot product between any given pair gives a sharp estimate of their respective RWK (see Fig. 1). **This guarantees linear scaling with respect to the size of the dataset $N_G$, as well as the number of nodes of each graph therein**. We derive novel exponential concentration bounds to prove that these Monte Carlo kernel estimates are accurate with high probability, and motivate our algorithmic design choices with theoretical discussion. The ability to approximate RWKs for arbitrary $(\mu_i)_{i=0}^{\infty}$ is especially important in settings where these coefficients are learned, unlocking scalable implicit graph kernel learning (see Sec. 5.4).

The corresponding code can be found here.

**Paper organization.** In Sec. 2, we review related work. In Secs 3.1, 3.2, and 3.3, we provide important mathematical preliminaries and discuss steps towards an $\mathcal{O}(N)$ algorithm. Sec. 3.4 gives the full GVoys method, whilst Sec. 4 provides theoretical guarantees. Empirical analysis, including speed comparisons with previous methods and downstream graph classification tasks, is presented in Sec. 5. We conclude in Sec. 6 and provide extra discussion and proofs in the Appendix.

## 2 RELATED WORK

The previous fastest methods for computing RWKs include Sylvester equation methods (SYLVs) (Du and Tong, 2018; Tian et al., 2024), conjugate gradient methods (CGs) (Nocedal and Wright, 1999; Zhang and Wang, 2012; Hager and Zhang, 2005), fixed point iteration methods (FPIs) (Bauschke et al., 2011; Kolesnikov and Oseledets, 2018) and spectral decomposition algorithms (SDs) (Pitarque et al., 1990; Jones et al., 2023; Miller, 2013). The first three approaches specifically consider the geometric walk kernel $\mu_i = \lambda^i$, such that $\mathrm{K}_{\mathrm{RWK}}(\mathrm{G}_1, \mathrm{G}_2) = \mathbf{v}^\top(\mathbf{I}_{N_1 \cdot N_2} - \lambda \mathbf{A}_{\mathrm{G}_1 \times \mathrm{G}_2})^{-1}\mathbf{w}$. The weaknesses of this choice of kernel were recently discussed by Sugiyama and Borgwardt (2015). One can take various approaches to solve this linear system. SYLVs recast the kernel as the solution to a generalized Sylvester equation, which takes $\mathcal{O}(N^3)$ time for both sparse and dense graphs. It requires that the adjacency matrix can be decomposed into one or two sums of Kronecker products, or else the the theoretical running time is unknown (Vishwanathan et al., 2006). SYLVs also require that $N_1 = N_2$ so the input graphs to have the same number of vertices, which is not realistic in most ML applications. Meanwhile, CGs directly solve the linear system, requiring time $O(m^2 k_{\mathrm{CG}})$ with $m$ the number of edges of the graph and $k_{\mathrm{CG}}$ the number of iterations of the CG-algorithm. Problematically, CG requires $O(m^2)$ memory, so its memory footprint is $O(N^2)$ even for sparse graphs with $m = \Theta(N)$. FPIs can be implemented in time $O(m^2 k_{\text{FPI}})$, where $k_{\text{FPI}}$ denotes for the number of iterations of the FPI algorithm. Detailed discussion thereof can be found in Vishwanathan et al. (2010, 2006); Kang et al. (2012); Kalofolias et al. (2021). Kang et al. (2012) provides an $O(N)$ algorithm, but only for the geometric RWK. On the other hand, spectral decomposition methods (SDs) are based on diagonalizing $\mathbf{A}_{\text{G}_1 \times \text{G}_2}$ and efficiently computing the Taylor series in this new basis. SDs work for general RWKs, but the time complexity is still $\mathcal{O}(N^3)$ (Thm. 4, Vishwanathan et al., 2010) and graphs must be unlabelled. Importantly, none of these variants is capable of $\mathcal{O}(N)$ unbiased approximation of general RWKs for labelled and unlabelled graphs.

# 3 GRAPH VOYAGERS (GVOYS)

## 3.1 Preliminaries: Graph Random Features

**Graph features.** Given an undirected graph G, consider the matrix $\mathbf{M}(\text{G}) = \sum_{i=0}^{\infty} \mu_i \mathbf{A}_{\text{G}}^i$, where $\mathbf{A}_{\text{G}} \coloneqq [\mathbb{I}[(i,j) \in \text{E}(\text{G})]]_{i,j=1}^{N}$ denotes the adjacency matrix of G and $(\mu_i)_{i=0}^{\infty}$ is a sequence of coefficients that provides convergence. For suitable instantiations of the sequence $(\mu_i)_{i=0}^{\infty}$, $\mathbf{M}(\text{G})$ is positive definite. Note that $\mathbf{M}(\text{G})$ is always symmetric. Therefore, its entries can be interpreted as evaluations of the kernel $\text{K}_{\text{G}} : \text{V} \times \text{V} \rightarrow \mathbb{R}$ defined on the set $\text{V}(\text{G})$ (rather than between multiple graphs). Let $(f_i)_{i=0}^{\infty}$ denote the discrete deconvolution of $(\mu_i)_{i=0}^{\infty}$, so that $\sum_{p=0}^{k} f_p f_{k-p} = \mu_k \ \forall \ k$. Up to a sign, this is unique. We refer to $f$ as the *modulation function*. It is a straightforward exercise to show that $\mathbf{M}(\text{G}) = \mathbf{\Phi}_{\text{G}} \mathbf{\Phi}_{\text{G}}^{\top}$ if $\mathbf{\Phi}_{\text{G}} \coloneqq \sum_{k=0}^{\infty} f_k \mathbf{A}_{\text{G}}^k \in \mathbb{R}^{N \times N}$ (Reid et al., 2024b). We call the rows of $\mathbf{\Phi}$ *graph features* $\{\phi_{\text{G}}(v_i)\}_{v_i \in \text{V}(\text{G})}$, so that $\mathbf{\Phi} = [\phi_{\text{G}}(v_i)]_{i=1}^{N}$ with $\phi_{\text{G}}(v_i) \in \mathbb{R}^N$. We have that $\text{K}_{\text{G}}(v_1, v_2) = \phi_{\text{G}}(v_1)^{\top} \phi_{\text{G}}(v_2)$ for all $v_1, v_2 \in \text{V}(\text{G})$. This factorization can also be conducted when $\mathbf{M}(\text{G})$ is not positive definite, but in this case some of the coefficients $f_i$ are complex.

**Graph <u>random</u> features.** One can approximate graph features $\{\phi_{\text{G}}(v_i)\}_{v_i \in \text{V}(\text{G})}$ by randomized low-dimensional or sparse estimates: graph *random* features (GRFs; Choromanski, 2023; Reid et al., 2024b). Intuitively, graph features depend on weighted powers of the adjacency matrix $\mathbf{A}$. The term $f_k \mathbf{A}^k$ simply counts the number of walks of length $k$ between every possible start and end vertex, weighted by the modulation function $f$. One can estimate this quantity by sampling simple random walks on G and measuring the empirical vertex occupations after $k$ timesteps. We use importance sampling since decaying $f$ means shorter walks contribute more to the sum. This allows us to compute a set of features $\{\widehat{\phi}_{\text{G}}(v_i)\}_{v_i \in \text{V}(\text{G})} \subset \mathbb{R}^r$, which are either sparse or satisfy $r \ll N$. More concretely, let $\Omega \coloneqq \left\{ (v_i)_{i=1}^{l} \,|\, v_i \in \text{V}(\text{G}), (v_i, v_{i+1}) \in \text{E}(\text{G}), l \in \mathbb{N} \right\}$ denote the set of *walks* on G, series of vertices connected by edges. Given a set of $m$ walks $\{\omega_k(v_i)\}_{k=1}^{m} \subset \Omega$ beginning at vertex $v_i$, we define the GRF as follows:

$$\widehat{\phi}(v_i)_{\text{G}} \coloneqq \frac{1}{m} \sum_{k=1}^{m} \sum_{\omega \text{ p.s. } \omega_k(v_i)} \frac{f_{\text{len}(\omega)}}{p(\omega)} \mathbf{e}_{\omega[-1]}. \quad (2)$$

Here, $\omega$ p.s. $\omega_k(v_i)$ means that $\omega$ is a *prefix subwalk* of $\omega_k(v_i)$, so $\omega = \omega_k(v_i)[: \text{len}(\omega)]$. $\text{len}(\omega)$ is the number of hops in the walk. $\omega[-1]$ is the last vertex of walk $\omega$, and $\mathbf{e}_k$ is the the unit vector for coordinate $k$. $p(\omega)$ is the probability of sampling the walk $\omega$. Crucially, GRFs satisfy $\text{K}_{\text{G}}(v_1, v_2) = \mathbb{E}[\widehat{\phi}_{\text{G}}(v_1)^{\top} \widehat{\phi}_{\text{G}}(v_2)]$ (Reid et al., 2024b). As written, Eq. 2 gives sparse $N$-dimensional features. The dimensionality can be reduced to $r < N$ whilst preserving unbiasedness of dot products by subsampling coordinates via *anchor points* (see Choromanski (2023)). Equipped with GRFs, one can write a randomized decomposition $\mathbf{M}(\text{G}) = \mathbb{E}[\mathbf{C}\mathbf{D}^{\top}]$, where $\mathbf{C}, \mathbf{D} = [\widehat{\phi}(v_i)]_{v_i \in \text{V}(\text{G})} \in \mathbb{R}^{N \times r}$ but with independent walks in each case.

**Time complexity of GRFs.** Each GRF $\widehat{\phi}_{\text{G}}(v_i)$ is obtained by simulating $m$ random walks on G, beginning at vertex $v_i$. It is standard to assume that they terminate with probability $p_{\text{halt}}$ at every timestep, whereupon the expected time complexity of the $(\mathbf{C}, \mathbf{D})$-decomposition is $O(Nm \frac{1}{p_{\text{halt}}})$. This is linear in the number of vertices.

## 3.2 Naive GRFs for an $O(N^2)$ Algorithm

If the product graph $\text{G}_1 \times \text{G}_2$ is instantiated in memory, GRFs can be straightforwardly applied to approximate $\mathbf{M}(\text{G}_1 \times \text{G}_2) \overset{\text{def}}{=} \sum_{i=0}^{\infty} \mu_i \mathbf{A}_{\text{G}_1 \times \text{G}_2}^i$ with a low-rank or sparse decomposition. Since $\text{K}_{\text{RWK}}(\text{G}_1, \text{G}_2) = \mathbf{v}^{\top} \mathbf{M}(\text{G}_1 \times \text{G}_2) \mathbf{w}$, we can write:

$$\text{K}_{\text{RWK}}(\text{G}_1, \text{G}_2) \overset{\mathbb{E}}{=} (\mathbf{v}^{\top} \mathbf{C})(\mathbf{D}^{\top} \mathbf{w}). \quad (3)$$

Here, $\mathbf{C}, \mathbf{D} \in \mathbb{R}^{N_1 N_2 \times r}$ are constructed from independent GRFs, taking $\mathbf{C} = [\widehat{\phi}_{\text{G}_1 \times \text{G}_2}(v_i)]_{v_i \in \text{V}(\text{G}_1 \times \text{G}_2)}$ and likewise for $\mathbf{D}$. The parentheses show the order of computation. Evaluating Eq. 3 incurs time complexity $O(N_1 N_2 r)$, which is quadratic if $N_1 = N_2 = N$. Naive application of GRFs already improves upon the time complexity of previously proposed algorithms for labelled and unlabelled graphs in Sec. 2, applicable to arbitrary choices of $(\mu_i)_{i=0}^{\infty}$.

**Limitations of the naive approach.** Whilst this approach provides a speedup, it still requires explicit computation of the product graph $\text{G}_1 \times \text{G}_2$ for every pair of inputs in order to sample random walks. Clearly, this must incur quadratic time complexity, even if the inputs are sparse. It also does not scale well to large datasets since the number of product graphs to be instantiated is quadratic in the number of graphs $N_{\mathrm{G}}$. It would be preferable to avoid this. In the following sections, we will see how this can be achieved by simulating *dependent* walks on $\mathrm{G}_1$ and $\mathrm{G}_2$, introducing extra random variables to sample walks on the product graph but without storing it in memory. This will allow us to estimate RWKs in $\mathcal{O}(N)$ time for sparse graphs. As a byproduct, we obtain novel low-dimensional embeddings for each graph (c.f. each graph vertex, which is achieved with GRFs), which unlocks $\mathcal{O}(N_{\mathrm{G}})$ linear time complexity scaling with respect to dataset size.

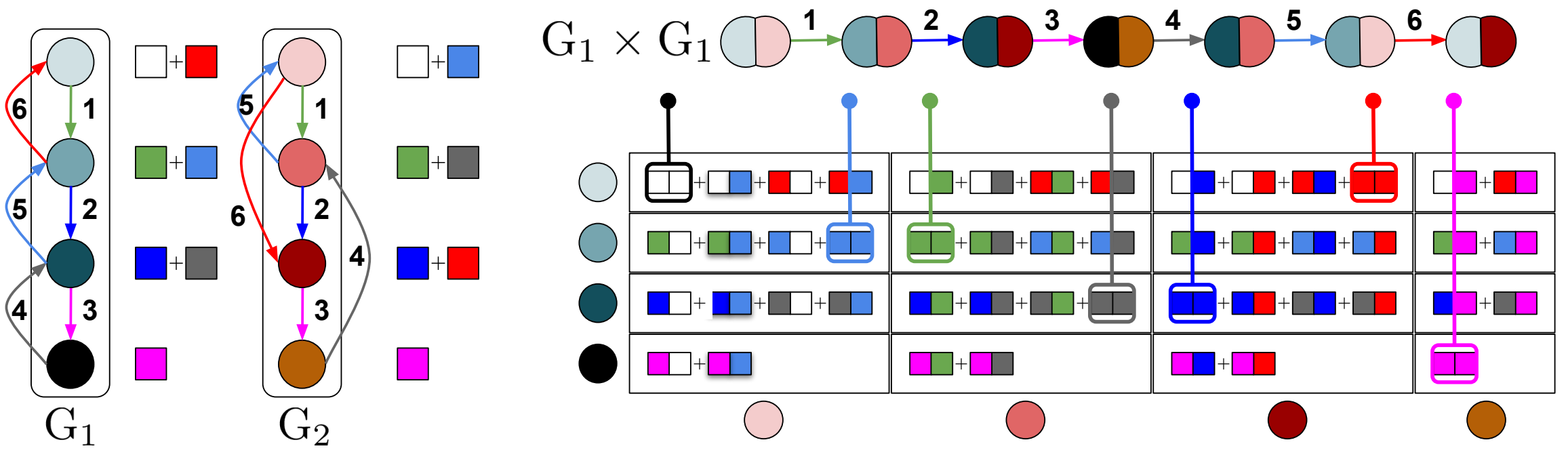

Figure 2: Schematic of the role of the random $g$-variables. **Left**: two graphs $\mathrm{G}_1, \mathrm{G}_2$ with random walks. The numbers show the hop order. At each timestep $i$, the walkers leave a 'deposit' at the corresponding graph node, modulated by a draw of a random Rademacher variable $g(i)$. This variable is represented by a coloured square: white, green, dark blue, pink, grey, light blue then red. **Lower right:** total loads deposited at each supervertex in the direct product graph $\mathrm{G}_1 \times \mathrm{G}_2$, of which there are $4 \times 4 = 16$ in total. Since the Rademacher draws are independent at different timesteps, $\mathbb{E}(g(i_1)g(i_2)) = \mathbb{I}(i_1 = i_2)$ so we only get a contribution (in expectation) when the colours of the squares match. Averaging, we filter deposits in each supervertex for $i_1 = i_2$. **Upper right:** retaining only these non-vanishing contributions where $i_1 = i_2$, we emulate the corresponding walk on the product graph, but without needing to instantiate it in memory explicitly.

### 3.3 Towards the Linear Algorithm

How can we approximate $\mathrm{K}_{\mathrm{RWK}}(\mathrm{G}_1, \mathrm{G}_2)$ without suffering the computational bottleneck of instantiating the product graph in memory? For unlabelled graphs, $\mathrm{K}_{\mathrm{RWK}}(\mathrm{G}_1, \mathrm{G}_2)$ can be written as:

$$\mathbf{v}^\top \left[ \left( \sum_{i=0}^{\infty} f_i \mathbf{A}^i_{\mathrm{G}_1 \times \mathrm{G}_2} \right) \left( \sum_{j=0}^{\infty} f_j \mathbf{A}^j_{\mathrm{G}_1 \times \mathrm{G}_2} \right) \right] \mathbf{w}. \tag{4}$$

Letting $\otimes$ denote the outer product,

$$\begin{aligned} \sum_{i=0}^{\infty} f_i \mathbf{A}^i_{\mathrm{G}_1 \times \mathrm{G}_2} &= \sum_{i=0}^{\infty} f_i \left( \mathbf{A}_{\mathrm{G}_1} \otimes \mathbf{A}_{\mathrm{G}_2} \right)^i \\ &= \sum_{i_1, i_2 = 0}^{\infty} \sqrt{f_{i_1}} \mathbf{A}^{i_1}_{\mathrm{G}_1} \otimes \sqrt{f_{i_2}} \mathbf{A}^{i_2}_{\mathrm{G}_2} \mathbb{I}(i_1 = i_2), \end{aligned} \tag{5}$$

where $\mathbb{I}(\cdot)$ is the indicator function that evaluates to 1 if its argument is true and 0 otherwise. This shows that the number of walks of a given length on $\mathrm{G}_1 \times \mathrm{G}_2$ is equal to the product of the number on $\mathrm{G}_1$ and $\mathrm{G}_2$ respectively. We already know how to efficiently approximate the sums $\sum_{i_1=0}^{\infty} \sqrt{f_{i_1}} \mathbf{A}^{i_1}_{\mathrm{G}_2}$ and $\sum_{i_2=0}^{\infty} \sqrt{f_{i_2}} \mathbf{A}^{i_2}_{\mathrm{G}_2}$ using GRFs (Eq. 2), but we must make a modification to filter out the cross terms where $i_1 \neq i_2$.

**Length random variables.** To incorporate the extra factor of $\mathbb{I}(i_1 = i_2)$, we propose to multiply the estimators of the $i$th powers $\mathbf{A}^i_{\mathrm{G}_1}$ and $\mathbf{A}^i_{\mathrm{G}_2}$ (corresponding to the $i$th hop on the respective graphs) by a shared random variable $g(i)$ with zero mean and unit variance. This is chosen so that $\mathbb{E}(g(i_1)g(i_2)) = \mathbb{I}(i_1 = i_2)$. We refer to these as *$g$-variables* (see Fig. 2). Concretely, we take:

$$\widehat{\phi}(v_i, g)_{\mathrm{G}} = \frac{1}{m} \sum_{k=1}^{m} \sum_{\omega \text{ p.s. } \omega_k(v_i)} \frac{\sqrt{f_{\mathrm{len}(\omega)}}}{p(\omega)} \mathbf{e}_{\omega[-1]} g(\mathrm{len}(\omega)), \tag{6}$$

for $\mathrm{G} \in \{\mathrm{G}_1, \mathrm{G}_2\}$. The random sequence $(g(l))_{l=0}^{\infty}$ must be shared between the (otherwise independent) walkers on the two graphs.

**What about labels?** If the graph vertices are also equipped with labels, we only include supervertices $(v_1, v_2) \in \mathrm{V}(\mathrm{G}_1) \times \mathrm{V}(\mathrm{G}_2)$ in the direct product graph when their labels match ($\mathcal{L}(v_1) = \mathcal{L}(v_2)$). To account for this, we make yet another modification to the estimator in Eq. 6 to remove contributions from walks $\omega_1$ on $\mathrm{G}_1$ and $\omega_2$ on $\mathrm{G}_2$ where the sequences of vertex labels differ, since in this case the composite walk on $\mathrm{G}_1 \times \mathrm{G}_2$ visits supervertices that do not exist. This can be achieved by introducing additional shared zero-mean and unit-variance *$z$-variables* $(z(l,i))_{l=1,\dots,N_L}^{i=0,1,\dots}$, where $N_L$ denotes the number of vertex labels. We take:

$$\begin{aligned} \widehat{\phi}(v_i, g, z)_{\mathrm{G}} = \frac{1}{m} \sum_{k=1}^{m} \sum_{\omega \text{ p.s. } \omega(v_i)_k} \frac{\sqrt{f_{\mathrm{len}(\omega)}}}{p(\omega)} \mathbf{e}_{\omega[-1]} \\ \cdot\, g(\mathrm{len}(\omega)) \prod_{l=0}^{\mathrm{len}(\omega)} z(\mathcal{L}(\omega[l]), l). \end{aligned} \tag{7}$$

These features will now successfully estimate contributions from walks on $\mathrm{G}_1 \times \mathrm{G}_2$, including with vertex labels, but without instantiating the product explicitly and thus with linear time complexity in $N$. Again, the dimensionality can be reduced to $r$ by subsampling anchor points without breaking unbiasedness.

**Putting it all together.** Define the matrices $\mathbf{C}_i = [\widehat{\phi}_{\mathrm{G}_i}(v_j)]_{v_j \in \mathrm{V}(\mathrm{G}_i)} \in \mathbb{R}^{N_i \times r_i}$ for $i \in \{1, 2\}$, constructed from $m$ random walks on $\mathrm{G}_i$ with shared $g$ and $z$ random variables to filter for walks with the same length and vertex label sequence. Construct $\mathbf{D}_i = [\widehat{\phi}_{\mathrm{G}_i}(v_j)]_{v_j \in \mathrm{V}(\mathrm{G}_i)} \in \mathbb{R}^{N_i \times r_i}$ completely analogously, but drawing a different set of walks and $g$ and $z$ independently. Making the popular assumption that $\mathbf{v} = \mathrm{flat}(\mathbf{v}_1 \otimes \mathbf{v}_2)$ and $\mathbf{w} = \mathrm{flat}(\mathbf{w}_1 \otimes \mathbf{w}_2)$, standard linear algebra for matrix outer products implies that

$$\mathrm{K}_{\mathrm{RWK}}(\mathrm{G}_1, \mathrm{G}_2) = \mathbb{E} \prod_{i=1}^{2} \left[ \left( \mathbf{v}_i^\top \mathbf{C}_i \right) \left( \mathbf{D}_i^\top \mathbf{w}_i \right) \right]. \quad (8)$$

Crucially, the time complexity of computing each of the curly brackets is $\mathcal{O}(N_i r_i)$, so is linear in the number of graph vertices. Moreover, the quantities in square brackets are scalars that only depend on $\mathrm{G}_1$ and $\mathrm{G}_2$ respectively, so taking $d_\mathrm{G}$ sets of $g$ and $z$ variables and draws of random walks (with the corresponding matrices $\mathbf{C}_i^{(k)}, \mathbf{D}_i^{(k)}$ for $k = 1, ..., d_\mathrm{G}$) and concatenating the results, we can construct $d_\mathrm{G}$-dimensional graph-level random features of the following form:

$$\phi(\mathrm{G}_i) := \frac{1}{\sqrt{d_\mathrm{G}}} [\left( \mathbf{v}_i^\top \mathbf{C}_i^{(k)} \right) \left( (\mathbf{D}_i^{(k)})^\top \mathbf{w}_i \right)]_{k=1}^{d_\mathrm{G}}. \quad (9)$$

These satisfy $\mathrm{K}_{\mathrm{RWK}}(\mathrm{G}_i, \mathrm{G}_j) = \mathbb{E} \left( \phi(\mathrm{G}_i)^\top \phi(\mathrm{G}_j) \right)$ for any pair of inputs, providing theoretically-grounded $d_\mathrm{G}$-dimensional representations of $\mathrm{G}_1$ and $\mathrm{G}_2$.

**Scaling with number of nodes, scaling with number of graphs.** Whilst the analysis above has focused on approximating the RWK with just two graphs, the graph-level random features from Eq. 9 can be constructed for *every* graph in a set $\{\mathrm{G}_i\}_{i=1}^{N_\mathrm{G}}$ (where $N_\mathrm{G}$ denotes the size of the dataset). The $g$ and $z$ variables must be cached and reused to compute every graph's feature. Let $\mathbf{K}_{\mathrm{RWK}} := [\mathrm{K}_{\mathrm{RWK}}(\mathrm{G}_i, \mathrm{G}_j)]_{i,j=1}^{N_\mathrm{G}} \in \mathbb{R}^{N_\mathrm{G} \times N_\mathrm{G}}$ denote the Gram matrix for the dataset. Define $\mathbf{\Phi} := [\phi(\mathrm{G}_i)]_{i=1}^{N_\mathrm{G}} \in \mathbb{R}^{N_\mathrm{G} \times d_\mathrm{G}}$, the *design matrix*. Then clearly $\mathbf{K}_{\mathrm{RWK}} = \mathbb{E} \left( \mathbf{\Phi} \mathbf{\Phi}^\top \right)$.

This low rank decomposition of the Gram matrix is important because it unlocks better scalability with respect to *the number of graphs* $N_\mathrm{G}$, in addition to their size $N$. For example, in analogy to the celebrated class of random Fourier features (RFFs; Rahimi and Recht, 2007), the decomposition permits $\mathcal{O}(N_\mathrm{G} d_\mathrm{G}^2)$ time complexity for computing the (approximate) posterior of a Gaussian process between graphs (Lázaro-Gredilla et al., 2010). The exact computation using $\mathbf{K}_{\mathrm{RWK}}$ explicitly is $\mathcal{O}(N_\mathrm{G}^3)$. To our knowledge, our algorithm is the first to guarantee linear scaling with respect to $N_\mathrm{G}$; previous methods focus on accelerating computation of $\mathrm{K}_{\mathrm{RWK}}(\mathrm{G}_1, \mathrm{G}_2)$, but are still required to do so explicitly for every pair of graphs in the dataset. Our algorithm enjoys linear scaling in *two* senses: with respect to the number of graph nodes, and with respect to the number of graphs.

**Remaining algorithmic choices.** We have presented a general method for unbiased approximation of RWKs in $\mathcal{O}(N)$ time; some flexibility still remains. For our experiments, we make the following design choices.

1. *Distributions of $g$ and $z$.* Any distribution with zero mean and unit variance will suffice for $g$ and $z$. We choose *Rademacher random variables*, taking values $\{\pm 1\}$ with equal probability. In Sec. 4, we prove that this is optimal for $g$ (Thm. 4.3).

2. *Sharing of $g$ and $z$ between walkers.* Eq. 6 assumes that all $m$ walkers on $\mathrm{G}_1$ and $\mathrm{G}_2$ share the same sequence of $g$ and $z$ variables. However, it is also possible to choose that the $k$th set of walks on $\mathrm{G}_1$ and $\mathrm{G}_2$ has their own independent sequence of variables: $(g(i,k))_{i=0,1,...}$ and $z(l,i,k)_{l=1,...,N_L}^{i=0,1,...}$ for $k \in \{1, ..., m\}$. One must modify the normalization to $\frac{1}{\sqrt{m}}$ to account for the fact that walkers with different $k$ no longer interact in expectation. We find that this reduces the kernel estimator variance compared to sharing $g$ since we take more samples. This is the choice we make in experiments.

3. *Random walk sampling.* Any reasonable choice of walk sampler $p(\omega)$ is possible since we divide by the known probability during importance sampling. Following convention, we consider *simple random walks* that choose one of their neighbours uniformly at random. We terminate with probability $p_{\mathrm{halt}}$ at every timestep, so that the lengths are geometrically distributed. For the $k$th set of walkers on different graphs, it is convenient to share the corresponding termination random variables $(t(l,k))_{l=0,1,...}$, whereupon we must normalize by $p(\omega) = \prod_{i=0}^{\mathrm{len}(\omega)-1} \frac{\sqrt{1-p_{\mathrm{halt}}}}{d_{\omega[i]}}$ with $d_{\omega[i]}$ the degree of the $i$th vertex of walk $\omega$.[1] Other sampling choices (e.g. independent termination between graphs) are also possible, provided the importance sampling weights are adjusted accordingly.

[1] Since the termination random variables are shared, this is not strictly the marginal probability of sampling walk $\omega$. It is chosen so that we get a factor of $(1 - p_{\mathrm{halt}})^{\mathrm{len}(\omega)}$ when we take the product in e.g. Eq. 5, which is equal to the joint probability under the sampling mechanism.

### 3.4 Graph Voyagers via Random Walks

Equipped with the discussion above, we are ready to present our novel GVoys algorithm. See Alg. 1 below.

**Algorithm 1** Sample $\{\mathbf{X_i}\}_{i=1}^{N_\mathrm{G}} \subset \mathbb{R}^{N_i \times r_i}$

**Input:** Vector of unweighted vertex degrees $\boldsymbol{d} \in \mathbb{R}^N$, modulation function $f_\mu : (\mathbb{N} \cup \{0\}) \to \mathbb{R}$, termination probability $p_\mathrm{halt} \in (0,1)$, number of random walks $m$, labelling function $\mathcal{L} : \mathrm{V}(\mathrm{G}_i) \to \{1, ..., N_L\}$.

**Output:** $\{\mathbf{X_i}\}_{i=1}^{N_\mathrm{G}}$ to estimate RWKs (with $\mathbf{X}_i \in \{\mathbf{C}_i, \mathbf{D}_i\}$)

Initialize $(g(l,w))_{l=0,1,...}^{w=0,1,...,m} \sim \mathrm{Rad.}(\pm 1)$
Initialize $(z(n,l,w))_{n=1,...,N_L}^{w=0,...,m,l=0,1,...} \sim \mathrm{Rad.}(\pm 1)$
Initialize $(t(l,w))_{l=0,1,...}^{w=0,1,...,m} \sim \mathrm{Unif}(0,1)$
**for** $\mathrm{G}_i$ in $\{\mathrm{G_i}\}_{i=1}^{N_\mathrm{G}}$ **do**
  Initialize $s[w][j] = \mathbf{0} \in \mathbb{R}^{N_i}$ for $w = 1, ..., m$, $j = 0, ..., r_i - 1$, sample anchors $a_0, ..., a_{r_i-1} \in \mathrm{V}(\mathrm{G}_i)$.
  **for** $w = 1, ..., m$ **do**
    **for** $k = 0, ..., N_i - 1$ **do**
      Initialize: `terminated` $\leftarrow$ False
      Initialize: `load` $\leftarrow 1$, `c` $\leftarrow k$, `l` $\leftarrow 0$
      `load` $\leftarrow$ `load` $\times z(L(k), 0, w)$
      **while** `not terminated` and $\exists_j$ `c` $= \mathbf{a}_j$ **do**
        deposit $= g(l,w) \times$ load $\times \sqrt{f_\mu(l)}$
        $s[w][j][\mathrm{k}] \leftarrow s[w][j][\mathrm{k}]$+deposit
        `terminated` $\leftarrow (t(l,w) < p_\mathrm{halt})$
        `n` $\leftarrow$ Unif [Neigh( `c` )]
        `load` $\leftarrow$ `load`$\times \frac{d[\mathtt{c}]}{\sqrt{1-p_\mathrm{halt}}}$
        `c` $\leftarrow$ `n`, `l` $\leftarrow$ `l`+1
        `load` $\leftarrow$ `load` $\times z(\mathcal{L}(c), l, w)$
      **end while**
    **end for**
  **end for**
  return: $\mathbf{X}_i = \sqrt{\frac{N_i}{m r_i}} (\sum_{w=1}^m s[w][j][k])^\top \in \mathbb{R}^{N_i \times r_i}$
**end for**

We introduced $(t(l,w))_{l=0,1,...,}^{w=0,1,...,m}$, an array of random variables sampled i.i.d. from $\mathrm{Unif}(0,1)$ which we use to decide whether the each walker terminates at every timestep. $\mathrm{Neigh}(v)$ denotes the set of neighbors of vertex $v$ of G. We remark that the exponential kernel takes $\mu_k = \frac{\lambda^k}{k!}$, whereupon $f_k = \frac{\lambda^k}{2^k k!}$. Meanwhile, the geometric kernel takes $\mu_k = \lambda^k$, whereupon $f_k = \frac{(2k-1)!!}{2^k k!}$. These are both included as special cases. For unlabelled graphs, simply set $N_L = 1$. The $\sqrt{N_i / r_i}$ normalization is included to preserve unbiasedness when we subsample anchor points; see Eq. 24 in App. A.2.

**Building features for multiple graphs.** To construct $\{\mathbf{C}_i\}_{i=1}^{N_\mathrm{G}}$ for the whole dataset, we need to use the same $g$, $z$ and $t$ variables for each graph. These are sampled once at the start of the algorithm and cached. This can even be before any graphs are seen (provided the number of labels is known in advance). We also remark that $\{\mathbf{C}_i\}_{i=1}^{N_\mathrm{G}}$ and $\{\mathbf{D}_i\}_{i=1}^{N_\mathrm{G}}$ are sampled as independent copies, so in practice we run Alg. 1 twice to get them both.

### 3.5 Time and Space Complexity

Theoretical analysis of Alg. 1 is given in Sec. 4. Here, we discuss its space and time complexity. The main data structure assembled in Alg. 1 is the three-dimensional tensor $s \in \mathbb{R}^{m \times r_i \times N_i}$. Thus, the space complexity is $O(mrN)$, where $m$ is the number of random walks, $r$ is the number of anchor vertices (sampled uniformly from V(G)) and N is the number of vertices of the graph. The expected time complexity is $O(\frac{1}{p_\mathrm{halt}} m \log(r) N + mrN)$, where $\frac{1}{p_\mathrm{halt}}$ is the average length of the random walk with stopping probability $p_\mathrm{halt}$ and the log factor corresponds to the efficient search in the anchor set ($\exists_j c = a_j$ check). For $m, r, \frac{1}{p_\mathrm{halt}} = O_N(1)$, the space and expected time complexity of Alg. 1 is $O(N)$. The same follows for approximation of $\mathrm{K_{RWK}}(\mathrm{G}_1, \mathrm{G}_2)$ with GVoys via Eq. 8.

**What if $\mathbf{r = N}$?** Even if the number of anchors is $N$, meaning all the vertices are anchor points and there is no dimensionality reduction of $\widehat{\phi}_\mathrm{G}(v_i)$, all the computations can be still conducted in expected space and time complexity $O(N)$. To see this, fix some $w^*$ corresponding to a given random walk. Rather than storing $s[w^*] \in \mathbb{R}^{r_i \times N_i}$ explicitly for $i = 1, 2$, we can instead record only the pairs indices $(j, k)$ where $s[w^*][j][k] > 0$ and the corresponding positive values $s[w^*][j][k]$, thereby storing it *sparsely*. Denote by $n_+^{w^*}$ the number of such pairs after execution of Alg. 1. Using this sparse representation of $s$, the space complexity of Alg. 1 is $\mathcal{S}_i = O(\sum_{w=1}^m n_+^w)$ and the time complexity is $\mathcal{T}_i = \mathcal{S}_i$. Therefore, $\mathrm{K_{RWK}}(\mathrm{G}_1, \mathrm{G}_2)$ can be computed in space $\mathcal{S}_1 + \mathcal{S}_2$ and time $\mathcal{T}_1 + \mathcal{T}_2$. Since $\mathbb{E}[n_+^{w^*}] = \frac{1}{p_\mathrm{halt}} N_i$, the computation of $\mathrm{K_{RWK}}(\mathrm{G}_1, \mathrm{G}_2)$ can be conducted in expected space and time $O\left(\frac{1}{p_\mathrm{halt}}(N_1 + N_2)m\right)$. This is linear with respect to the number of nodes.

**Block-GVoys.** One can also exploit the structure of Eq. 9 to control the memory footprint of GVoys. Supposing we budget to simulate $m$ random walks per node in total, we can partition them into $\lfloor m/d_\mathrm{G} \rfloor$ blocks, where $d_\mathrm{G}$ is the size of the latent representation. Each block is used to sample $\{\mathbf{C}_i, \mathbf{D}_i\}_{i=1}^{N_\mathrm{G}}$ via Alg. 1. These are used to compute the next (scalar) entries of $\{\phi(\mathrm{G}_i)\}_{i=1}^{N_\mathrm{G}}$, and are then deleted from memory. We do not need to store $\{\mathbf{C}_i, \mathbf{D}_i\}_{i=1}^{N_\mathrm{G}}$ for every block of random walkers simultaneously – just the graph-level feature vectors, which are very compact (a set of $N_\mathrm{G}$ $d_\mathrm{G}$-dimensional vectors). In this way, we can tune $d_\mathrm{G}$ to reduce memory requirements as needed. We refer to this variant as *block-GVoys*.

## 4 THEORETICAL ANALYSIS

Here, we provide rigorous theoretical guarantees for our GVoys method (Alg. 1). We begin with our central

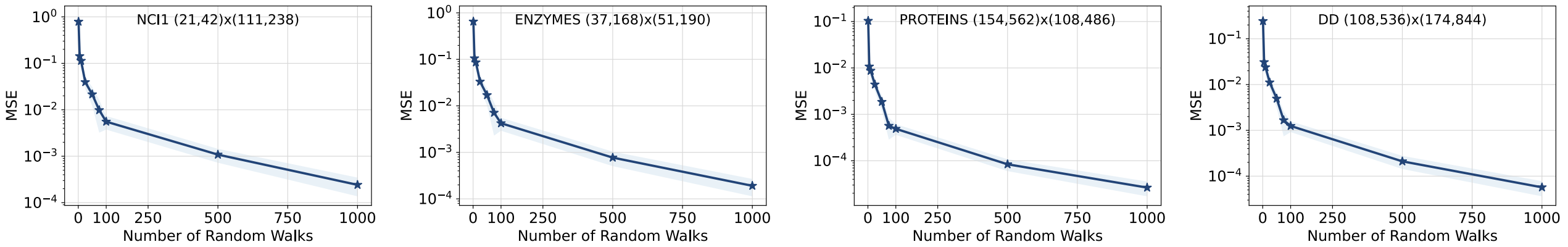

Figure 3: GVoy average kernel approximation error on graphs from TUDataset (Morris et al., 2020), plotted as a function of the number of sampled random walks $m$. The pair of tuples next to each dataset name corresponds to the number of vertices and edges of the respective graphs. Shaded regions represent standard errors over 10 runs.

claim that the approximation is unbiased.

**Theorem 4.1** (GVoys are unbiased)**.** *Supposing matrices* $\mathbf{C}_{1,2}$*,* $\mathbf{D}_{1,2} \in \mathbb{R}^{N_{1,2} \times r_{1,2}}$ *are sampled according to Alg. 1, the estimator*

$$\widehat{\mathrm{K}}_{\mathrm{RWK}}(\mathrm{G}_1, \mathrm{G}_2) = (\mathbf{v}_1^\top \mathbf{C}_1 \mathbf{D}_1^\top \mathbf{w}_1)(\mathbf{v}_2^\top \mathbf{C}_2 \mathbf{D}_2^\top \mathbf{w}_2) \quad (10)$$

*provides an unbiased estimate of the RWK, so that* $\mathrm{K}_{\mathrm{RWK}}(\mathrm{G}_1, \mathrm{G}_2) = \mathbb{E}(\widehat{\mathrm{K}}_{\mathrm{RWK}}(\mathrm{G}_1, \mathrm{G}_2))$.

*Proof sketch.* Sec. 3.3 outlines the motivation for Alg. 1, generalizing ideas from GRFs by incorporating extra random variables to emulate walks on $\mathrm{G}_1 \times \mathrm{G}_2$ without explicitly instantiating it. We also rely on standard linear algebra identities for operations with matrix outer products. App. A.1 gives full details. □

GVoys provide sharp estimates of graph kernels. Under mild assumptions on $\mathrm{G}_{1,2}$, we are able to provide exponential concentration bounds.

**Theorem 4.2** (GVoys give sharp kernel estimates)**.** *Let* $c(\mathrm{G}) := \sum_{l=0}^{\infty} \sqrt{|f_l|} \left( \frac{\max_{v \in \mathrm{G}} d_v}{\sqrt{1 - p_{halt}}} \right)^l \in \mathbb{R}$*. Suppose that* $c(\mathrm{G}_1)$ *and* $c(\mathrm{G}_2)$ *are finite. Further define*

$$k(m, \mathrm{G}_1, \mathrm{G}_2) := 2 \left( \frac{4(2m-1)}{m^2} + \frac{4(2m-1)^2}{m^4} \right) \cdot c(\mathrm{G}_1)^2 c(\mathrm{G}_2)^2 \|\mathbf{v}_1\|_1 \|\mathbf{w}_1\|_1 \|\mathbf{v}_2\|_1 \|\mathbf{w}_2\|_1, \quad (11)$$

*where* $m$ *is the number of random walks. Provided the* $g$ *and* $z$ *variables are* shared *across the* $m$ *walkers, conditioned on a particular draw of* $g$ *and* $z$*,*

$$\Pr\left( |\widehat{\mathrm{K}}(\mathrm{G}_1, \mathrm{G}_2 | g, z)| - \mathbb{E}(\widehat{\mathrm{K}}(\mathrm{G}_1, \mathrm{G}_2 | g, z))| \geq \epsilon \right) \leq 2 \exp\left( -2 \frac{\epsilon^2}{m k(m, \mathrm{G}_1, \mathrm{G}_2)^2} \right). \quad (12)$$

*Proof sketch.* Provided $c(\mathrm{G}_{\{1,2\}})$ is finite, we can bound the $L_1$ norm of the GRFs used to construct the rows of $\mathbf{C}_{\{1,2\}}$ and $\mathbf{D}_{\{1,2\}}$. This enables us to bound the change in the kernel estimator $\widehat{\mathrm{K}}(\mathrm{G}_1, \mathrm{G}_2)$ if one of the $m$ walks sampled per node changes. To apply McDiarmid's inequality, we need independence between the $m$ draws of random variables. This is not the case since $g$ and $z$ are shared, but we do have *conditional* independence given a particular draw of $g, z$. Applying the inequality, the result follows. See App. A.2. □

It is remarkable that the only dependence on the number of nodes $N$ is via the $L_1$ norm of the vectors, $\|\mathbf{v}_1\|_1, \|\mathbf{w}_1\|_1, \|\mathbf{v}_2\|_1, \|\mathbf{w}_2\|_1$. Fixing these as the graphs grow is a natural choice, in which case the bound becomes *independent of graph size.* We also remark that removing the conditioning on the draw of random variables $g, z$ is an important open problem. Relaxing it is left to future work, and will likely require different proof techniques to the bounded difference McDiarmid approach currently adopted.

We also noted in Sec. 3.3 that any distribution for $g$ with zero mean and unit variance is sufficient for unbiased estimation of RWKs, but that the Rademacher distribution is best. This is formalized as follows.

**Theorem 4.3** (Rademacher is optimal)**.** *Among all possible* $g$ *for which the estimator is unbiased, the* Rademacher distribution, $g(l, w) \in \{\pm 1\}$ *with equal probability, minimizes the variance of the RWK estimator* $\widehat{\mathrm{K}}(\mathrm{G}_1, \mathrm{G}_2)$ *in the GVoy algorithm.*

*Proof sketch.* We simulate random walks on $\mathrm{G}_1 \times \mathrm{G}_2$ by simulating random walks on $\mathrm{G}_1$ and $\mathrm{G}_2$ separately, also multiplying by the shared random variables $g$ so that (in expectation) we only include contributions where the lengths are equal. Writing out the expression for the expectation of the squared kernel estimator, the only surviving nontrivial terms depend on $\mathbb{E}(g(l, w)^4)$, the *kurtosis* of $g$. Since $\mathbb{E}(g(l, w)^4) \geq \mathbb{E}(g(l, w)^2)^2 = 1$, this is minimised by the Rademacher distribution which trivially has kurtosis 1. See App. A.3. □

## 5 EXPERIMENTS

Here, we answer the following questions: **(1)** How accurately can GVoys approximate RWKs? **(2)** How much faster are GVoys compared to other methods? **(3)** How do GVoys perform in downstream tasks? **(4)** Are there benefits to approximating general RWKs rather than just specific instantiations?

Table 1: Graph classification accuracies on the Benchmark TUDataset (Morris et al., 2020). The exact RWK values are from Nikolentzos et al. (2021). 'GVoy-L' denotes the labelled variant of our algorithm, whereas plain 'GVoy' takes $N_L = 1$.

| METHOD | MUTAG | ENZYMES | NCI1 | PTC-MR | D&D | PROTEINS | AIDS |
|---|---|---|---|---|---|---|---|
| Exact RWK | 81.4 (±8.9) | 16.7 (±1.8) | TIMEOUT | 54.4 (±9.8) | OOM | 69.5 (±5.1) | 79.0 (±2.3) |
| GVoy | 83.6 (±5.9) | **20.0** (±3.1) | 63.5 (±1.6) | **61.6** (±0.9) | **74.28** (±3.4) | **71.6** (±3.7) | **97.8** (±0.8) |
| GVoy-L | **84.1** (±6.6) | 19.9 (±4.2) | **64.2** (±1.7) | 55.8 (±5.9) | 73.9 (±3.4) | **71.6** (±3.7) | **97.8** (±0.8) |

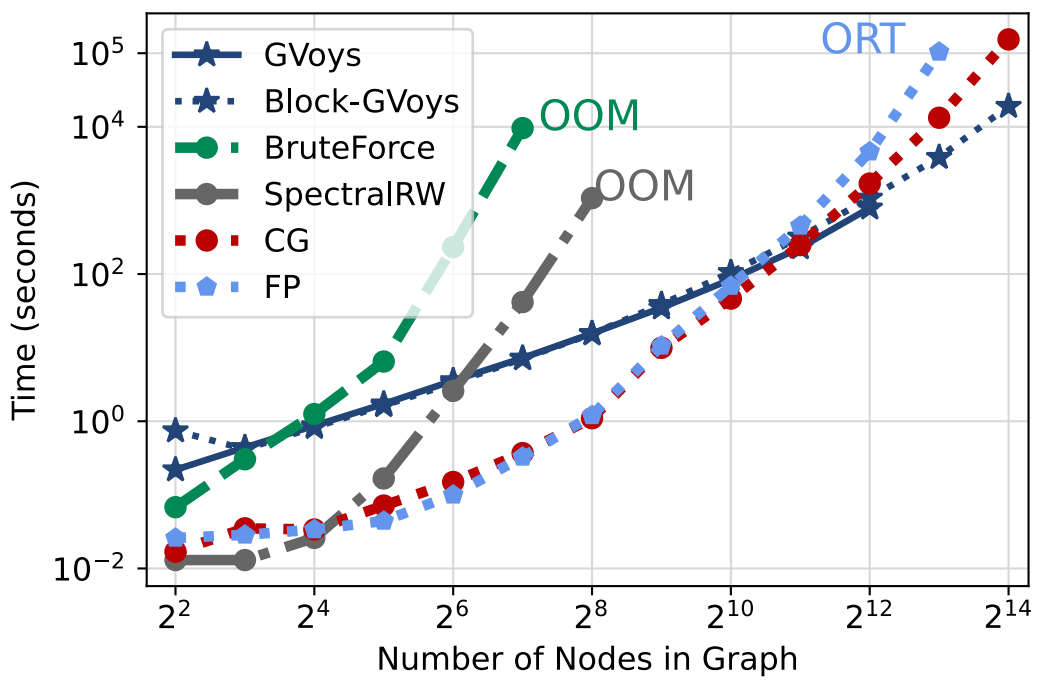

Figure 4: Comparison of GVoys runtime with brute force baseline ('BruteForce') and previous efficient methods, with increasing numbers of vertices $N$. We generate datasets of $N_{\text{G}} = 10$ Erdős-Rényi graphs with $N$ vertices each and edge probability $p = 0.1$. OOM and ORT mean 'out of memory' (32 GB) and 'out of runtime' (24 hours), respectively.

### 5.1 Quality Analysis

First we show that GVoys converge to the *exact* RWK as the number of walks increases. Fig. 3 illustrates the empirical mean squared error (MSE) of GVoys as a function of the number of random walks for graphs of varying sizes and sparsity levels, taken from the TUDatasets (Morris et al., 2020). As expected, the average approximation error drops as the number of walkers $m$ grows. Additional details and results are presented in Appendices B.1 and C.1.

### 5.2 Speed Analysis

Next, we compare GVoys with other efficient algorithms for computing the RWK: namely, spectral decomposition, conjugate gradient and fixed point iteration methods. Fig. 4 shows the results. As the graphs get larger, all methods take longer. Since our method is $\mathcal{O}(N)$, it becomes **faster than all baselines** for graphs with more than about $2^{10}$ nodes. For graphs with $2^{13}$ nodes, GVoys is **27**× faster than fixed-point iterations. For graphs with $2^{14}$ nodes, our algorithm is **8**× faster than conjugate gradient methods, while other baselines either run out of memory or exceed the allotted runtime. See App. B.2 for full details. In App. C.2, we include ablation results for the number of random walks; see Fig. 6. Even when sampling up to 1000 walks per node for high kernel approximation quality, our method is nearly **10**× faster than brute-force computation.

### 5.3 Downstream Task: Graph Classification

Now we apply GVoys to graph classification, comparing it to the baseline results from Nikolentzos et al. (2021) on 7 labeled graphs. These datasets vary in size from **200** to **4K** graphs. They consist of graphs with as few as 2 vertices and 2 edges to as many as **300** vertices and **1K** edges (see Table 2). We follow the same setup as in Nikolentzos et al. (2021) and report the **10-fold** cross validation results in Table 1. GVoys not only match the performance of the exact RWK, but in fact *surpass it* on all datasets considered. We posit that its stochastic nature makes it more robust to noise and outliers. We also note that the labelled GVoys variant does not always do better than the unlabelled version, likely because introducing extra $z$-variables increases the kernel estimator variance for a fixed number of walkers. Additional details are reported in App. B.3.

### 5.4 Learning Random Walk Kernels

Finally, we provide evidence that *graph kernel learning* via $(\mu_i)_{i=0}^{\infty}$ may be beneficial in downstream tasks, providing extra motivation for developing algorithms that can approximate general RWKs rather than just special cases. We truncate Eq. 1 at $n = 8$ terms and learn the Taylor coefficients end-to-end by minimizing the loss on a downstream classification task. On MUTAG, we obtain accuracy $88.3 \pm 3.1$, an **8.5**% relative improvement over the geometric RWK baseline (which achieves $81.4 \pm 8.9$). It is intuitive that the geometric RWK may not be the best choice for $(\mu_i)_{i=0}^{\infty}$ (Sugiyama and Borgwardt, 2015). GVoys permit scalable, implicit learning of a better kernel in linear time. See App. B.4 for details and App. C.3 for additional experiments.

## 6 CONCLUSION

We have introduced the first linear time and space complexity algorithms for unbiased approximation of general random walk graph kernels for sparse graphs, able to scale to $> 2^{14}$ nodes. In doing so, we have developed novel, theoretically-motivated graph embeddings (random features) which unlock better scalability with respect to the number of graphs in the dataset. We have provided detailed theoretical analysis and empirical evaluation, showcasing improved scalability and performance compared to spectral decomposition, conjugate gradient and fixed point iteration approaches.

# 7 ADDITIONAL DISCUSSION

## 7.1 General Comments

Our theoretical results show that asymptotically, for large enough graphs, GVOYs provide the fastest methods for RWK computation and this is confirmed in Fig. 4. Furthermore, the validity of the performed theoretical analysis is also illustrated by the nearly-constant slope of the GVOY-curve in Fig. 4.

Our presented algorithm is unbiased. This is the case for all unweighted undirected graphs, labeled or unlabeled, not only sparse (which are the chief focus of this paper). Instead, taking a biased algorithm would degrade approximation quality, since the MSE of the estimator would not go to zero as the number of random walks goes to infinity. Nevertheless, the impact this would have on the performance in downstream tasks is a more subtle question – depending on the effectiveness of the target RWK kernel for the particular task at hand, it is possible that a biased kernel (which can be interpreted as unbiased estimation of a different kernel than the original target) would be better.

## 7.2 Time Complexity: Deeper Dive

The time complexity of our method is $\mathcal{O}(\frac{Nm}{p})$ so if one chooses to increase the number of walkers $m$ with $N$, then the true time complexity will be superlinear in graph size. Our central claim is that presented algorithm is **the first method that provides an unbiased estimate of the graph kernel in $\mathcal{O}(N)$ time**. However, for the following reasons, we expect the number of walkers needed for a given kernel approximation error to scale sublinearly with the number of graph nodes:

- The bound in Eq. 12 does not depend on $N$, so we can choose the number of walkers independently of $N$ and still guarantee a good kernel estimate with high probability, conditioned on a given draw of $g$ and $z$ (since we condition on fixed $g$ and $z$, this is not yet quite sufficient to conclude that we can choose totally independently of, which is left as future work).
- There is strong evidence in the literature that, using GRFs with fixed $m$, the relative kernel approximation error increases slowly with (see for instance: Fig. 8, p.17 in Reid et al. (2024b)). We also confirmed this empirically for GVOY-s. We took the class of path-graphs, and fixed a small number of random walks. We then compared the predictions of our method with the exact kernel values on a wide range of graph sizes. The error is $< 0.003$.

In general, the optimal choice of $m$ depends on both: the graph structure (e.g. its mixing properties) as well as the particular structure of the RWK-kernel (in particular the $\mu$-sequence, e.g how fast it decreases, whether it has the constant number of nonzero terms of not, etc.).

Theorem 4.2 can be used to deduce the minimum number of walkers needed to achieve a sharp kernel estimate with high probability (conditioned on draws of $g$ and $z$). Given some desired quality $\epsilon$ and failure tolerance $\delta$, one can solve the bound for $m$.

Finally, note that even though brute-force computations for RWKs takes cubic time, there exist classes of graph kernels used in machine learning that support brute-force sub-cubic computation.

## 7.3 New Graph Embeddings

One of the biggest contributions of this paper, and the by-product of our algorithm for approximating RWKs, is the new class of graphs' latent embeddings. What is unique about it is that it provides strong theoretical guarantees since dot-products of those embeddings correspond to well-defined similarity measures between graphs (namely, general RWKs). Furthermore, those embeddings can still be learnable as components of deep neural network systems (coefficients $\mu$, vectors $v$, $w$). Learning those corresponds to training particular instantiations of the general RWKs.

Those graph embeddings can be leveraged also in algorithms operating on graph with feature-vectors in nodes. One way of doing it is via late fusion where the embedding of the graph obtained by applying regular Graph Neural Network operating on graphs with feature-vectors in nodes is concatenated with the GVOY-embedding (that takes into account only the graph structure rather than internal node representations) and processed by the final MLP module.

## Checklist

1. For all models and algorithms presented, check if you include:
   (a) A clear description of the mathematical setting, assumptions, algorithm, and/or model. [**Yes**/No/Not Applicable]. In section 3 and 4, we detail the problem description and our algorithm.
   (b) An analysis of the properties and complexity (time, space, sample size) of any algorithm. [**Yes**/No/Not Applicable]. Detailed description is provided in section 3, 4 and A.
   (c) (Optional) Anonymized source code, with specification of all dependencies, including external libraries. [Yes/**No**/Not Applicable]
2. For any theoretical claim, check if you include:
   (a) Statements of the full set of assumptions of all theoretical results. [**Yes**/No/Not Applicable]. The statements of all our main results are in Section 4.
   (b) Complete proofs of all theoretical results. [**Yes**/No/Not Applicable]. The complete proofs can be found in Section A.
   (c) Clear explanations of any assumptions. [**Yes**/No/Not Applicable]. Detailed explanations of all assumptions are presented in section 3 and 4.
3. For all figures and tables that present empirical results, check if you include:
   (a) The code, data, and instructions needed to reproduce the main experimental results (either in the supplemental material or as a URL). [**Yes**/No/Not Applicable]. The data is downloaded from TUDatasets. The baseline methods are run using the Grakel library.
   (b) All the training details (e.g., data splits, hyperparameters, how they were chosen). [**Yes**/No/Not Applicable]. All the experimental details is provided in Section B.
   (c) A clear definition of the specific measure or statistics and error bars (e.g., with respect to the random seed after running experiments multiple times). [**Yes**/No/Not Applicable]

(d) A description of the computing infrastructure used. (e.g., type of GPUs, internal cluster, or cloud provider). [**Yes**/No/Not Applicable]. This information is provided in Section B.

4. If you are using existing assets (e.g., code, data, models) or curating/releasing new assets, check if you include:

   (a) Citations of the creator If your work uses existing assets. [**Yes**/No/Not Applicable]. We cite all relevant works in the bibliography.

   (b) The license information of the assets, if applicable. [Yes/No/**Not Applicable**]. We use open source datasets.

   (c) New assets either in the supplemental material or as a URL, if applicable. [Yes/No/**Not Applicable**].

   (d) Information about consent from data providers/curators. [Yes/No/**Not Applicable**]. We only use open source benchmark datasets.

   (e) Discussion of sensible content if applicable, e.g., personally identifiable information or offensive content. [Yes/No/**Not Applicable**]

5. If you used crowdsourcing or conducted research with human subjects, check if you include:

   (a) The full text of instructions given to participants and screenshots. [Yes/No/**Not Applicable**]

   (b) Descriptions of potential participant risks, with links to Institutional Review Board (IRB) approvals if applicable. [Yes/No/**Not Applicable**]

   (c) The estimated hourly wage paid to participants and the total amount spent on participant compensation. [Yes/No/**Not Applicable**]

# Appendices: Optimal Time Complexity Algorithms for Computing General Random Walk Graph Kernels on Sparse Graphs

## A THEORY

In this appendix, we prove all theoretical claims in the main text.

### A.1 Proof of Thm. 4.1: GVoys are unbiased

Let us begin by proving Thm. 4.1: that GVoys provide an unbiased estimate of RWKs, so

$$\mathrm{K}_{\mathrm{RWK}}(\mathrm{G}_1,\mathrm{G}_2)=\mathbb{E}\left[(\mathbf{v}_1^\top\mathbf{C}_1\mathbf{D}_1^\top\mathbf{w}_1)(\mathbf{v}_2^\top\mathbf{C}_2\mathbf{D}_2^\top\mathbf{w}_2)\right] \tag{13}$$

with matrices $\mathbf{C}_1,\mathbf{C}_2,\mathbf{D}_1,\mathbf{D}_2$ computed using Alg. 1. Our discussion will follow the outline in Sec. 3.3 but with extra technical details.

*Proof.* Initially, consider unlabelled graphs. In Eqs 4 and 5, we found that

$$\mathrm{K}_{\mathrm{RWK}}(\mathrm{G}_1,\mathrm{G}_2)=\mathbf{v}^\top\left[\sum_{i=0}^{\infty}\mu_i\mathbf{A}^i_{\mathrm{G}_1\times\mathrm{G}_2}\right]\mathbf{w}=\mathbf{v}^\top\left[\left(\sum_{i=0}^{\infty}f_i\mathbf{A}^i_{\mathrm{G}_1\times\mathrm{G}_2}\right)\left(\sum_{j=0}^{\infty}f_j\mathbf{A}^j_{\mathrm{G}_1\times\mathrm{G}_2}\right)^\top\right]\mathbf{w} \tag{14}$$

if $\sum_{p=0}^{k}f_pf_{k-p}=\mu_k\ \forall\ k$. We also noted that

$$\sum_{i=0}^{\infty}f_i\mathbf{A}^i_{\mathrm{G}_1\times\mathrm{G}_2}=\sum_{i_1,i_2=0}^{\infty}\sqrt{f_{i_1}}\mathbf{A}^{i_1}_{\mathrm{G}_1}\otimes\sqrt{f_{i_2}}\mathbf{A}^{i_2}_{\mathrm{G}_2}\mathbb{I}(i_1=i_2), \tag{15}$$

whereupon standard identities for linear algebra with outer products imply that

$$\begin{aligned}\mathrm{K}_{\mathrm{RWK}}(\mathrm{G}_1,\mathrm{G}_2)=&\left[\mathbf{v}_1^\top\left(\sum_{i_1=0}^{\infty}\sqrt{f_{i_1}}\mathbf{A}^{i_1}_{\mathrm{G}_1}\right)\left(\sum_{j_1=0}^{\infty}\sqrt{f_{j_1}}\mathbf{A}^{j_1}_{\mathrm{G}_1}\right)^\top\mathbf{w}_1\right]\\&\cdot\left[\mathbf{v}_2^\top\left(\sum_{i_2=0}^{\infty}\sqrt{f_{i_2}}\mathbf{A}^{i_2}_{\mathrm{G}_2}\right)\left(\sum_{j_2=0}^{\infty}\sqrt{f_{j_2}}\mathbf{A}^{j_2}_{\mathrm{G}_2}\right)^\top\mathbf{w}_2\right]\mathbb{I}(i_1=i_2)\mathbb{I}(j_1=j_2).\end{aligned} \tag{16}$$

We can estimate $\sum_{i=0}^{\infty}\sqrt{f_i}\mathbf{A}^i_{\mathrm{G}}$ using random walks on G. Consider the GRF (Reid et al., 2024b),

$$\widehat{\phi}(v_i)_{\mathrm{G}}\coloneqq\frac{1}{m}\sum_{k=1}^{m}\sum_{\omega\text{ p.s. }\omega(v_i)_k}\frac{\sqrt{f_{\mathrm{len}(\omega)}}}{p(\omega)}\mathbf{e}_{\omega[-1]}. \tag{17}$$

Then note that

$$\begin{aligned}\mathbb{E}(\widehat{\phi}(v_i)_{\mathrm{G}})_q=\mathbb{E}\left(\sum_{\omega\text{ sampled}}\frac{\sqrt{f_{\mathrm{len}(\omega)}}}{p(\omega)}\mathbb{I}(\text{ends at node }q)\right)&=\sum_{\omega\text{ between }i\text{ and }q}\frac{\sqrt{f_{\mathrm{len}(\omega)}}}{p(\omega)}\mathbb{E}(\mathbb{I}(\omega\text{ is sampled}))\\&=\sum_{\omega\text{ between }i\text{ and }q}\sqrt{f_{\mathrm{len}(\omega)}}=\left(\sum_{k=0}^{\infty}\sqrt{f_k}\mathbf{A}^k_{\mathrm{G}}\right)_{iq},\end{aligned} \tag{18}$$

so $[\widehat{\phi}(v_i)_{\mathrm{G}}]_{v_i=1}^{N}\in\mathbb{R}^{N\times N}$ gives an unbiased estimate of $\sum_{k=0}^{\infty}\sqrt{f_k}\mathbf{A}^k_{\mathrm{G}}$.

Eq. 17 shows that GRFs work by simulating random walks $\{\omega(v_i)_k\}_{k=1}^N$ out of each node $v_i$, depositing 'load' $\frac{\sqrt{f_{\text{len}(\omega)}}}{p(\omega)}$ at each timestep at the coordinate corresponding to the walker's present location. As discussed in Sec. 3.3, the extra factor of $\mathbb{I}(i_1 = i_2)$ is obtained in expectation by multiplying the randomised estimates of $\mathbf{A}^i_{\text{G}_1}$ and $\mathbf{A}^i_{\text{G}_2}$ by Rademacher random variables $(g(i))_{i=0,1,...}$, This is achieved by modifying the GRFs computation to

$$\widehat{\phi}(v_i, g)_{\text{G}} = \frac{1}{m}\sum_{k=1}^{m}\sum_{\omega \text{ p.s. } \omega(v_i)_k} \frac{\sqrt{f_{\text{len}(\omega)}}}{p(\omega)}\mathbf{e}_{\omega[-1]}g(\text{len}(\omega)), \tag{19}$$

so that the deposit for the $i$th power is modulated by $g(i)$ (as in Eq. 6 of the main text). This works because the $i$th hop corresponds to the Monte Carlo estimate of the $i$th power of the adjacency matrix. Supposing we draw $m$ independent sequences $(g(l,w))_{l=0,1,...}^{w=0,1,...,m} \sim \text{Rad.}(\pm 1)$, one for each of the random walks, we must change the normalisation $\frac{1}{m} \to \frac{1}{\sqrt{m}}$ since $w_1 \neq w_2$ terms will no longer contribute to the approximation of the outer products $\mathbf{A}^i_{\text{G}_1\times\text{G}_2}$. Hence, we instead take

$$\widehat{\phi}(v_i, (g))_{\text{G}} = \frac{1}{\sqrt{m}}\sum_{k=1}^{m}\sum_{\omega \text{ p.s. } \omega(v_i)_k} \frac{\sqrt{f_{\text{len}(\omega)}}}{p(\omega)}\mathbf{e}_{\omega[-1]}g(\text{len}(\omega), k). \tag{20}$$

**Walk sampler.** As discussed in the main text, we are free to choose our walk sampler and $p(\omega)$. We take simple random walks that choose a neighbour uniformly at every timestep. If the termination random variables are shared, the probability of sampling a walk of length $\text{len}(\omega)$ in the product graph is trivially $(1-p)^{\text{len}(\omega)}$. To ensure we normalise the corresponding contributions appropriately, we can let $p(\omega) = \sqrt{1-p}^{\text{len}(\omega)} \prod_{v\in\omega[:-1]} \frac{1}{d_v}$. Note that this will *not* give the correct importance sampling weight when $i_1 \neq i_2$, but this is unimportant since these contributions vanish in expectation (by design with $g$); it *will* give the right normalisation for the $i_1 = i_2$ terms. To summarise, we can take

$$\widehat{\phi}(v_i, (g))_{\text{G}} = \frac{1}{\sqrt{m}}\sum_{k=1}^{m}\sum_{\omega \text{ p.s. } \omega(v_i)_k} \sqrt{f_{\text{len}(\omega)}}\mathbf{e}_{\omega[-1]}g(\text{len}(\omega), k)\prod_{v\in\omega[:-1]}\frac{d_v}{\sqrt{1-p_{\text{halt}}}}, \tag{21}$$

which codifies Alg. 1 as an equation. Crucially, we have that

$$\sum_{i=0}^{\infty} f_i \mathbf{A}^i_{\text{G}_1\times\text{G}_2} = \mathbb{E}\left(\mathbf{C}_1 \otimes \mathbf{C}_2\right) \quad \text{with} \quad \mathbf{C}_{1,2} \coloneqq \left[\widehat{\phi}(v_i, (g))_{\text{G}_{1,2}}\right]_{i=1}^N. \tag{22}$$

Constructing $\mathbf{D}_{1,2}$ analogously with independent copies of the random variables and plugging into Eq. 16, unbiasedness of $\widehat{\text{K}}_{\text{RWK}}(\text{G}_1, \text{G}_2)$ immediately follows.

**Dimensionality reduction and anchor points.** As written, Eq. 17 gives features $\widehat{\phi}(v_i) \in \mathbb{R}^N$ that satisfy

$$\mathbb{E}(\widehat{\phi}(v_i)^\top \widehat{\phi}(v_j)') = \sum_{q=1}^{N}\mathbb{E}(\widehat{\phi}(v_i)_q^\top \widehat{\phi}(v_j)'_q) = \left[\left(\sum_{i_1=0}^{\infty}\sqrt{f_{i_1}}\mathbf{A}^{i_1}_{\text{G}_1}\right)\left(\sum_{j_1=0}^{\infty}\sqrt{f_{j_1}}\mathbf{A}^{j_1}_{\text{G}_1}\right)^\top\right]_{ij} \quad \forall \quad v_i, v_j \in \text{V}(\text{G}). \tag{23}$$

Suppose we randomly sample $r \leq N$ coordinates without replacement. Denote the sampled set by $\mathcal{S}_r \subset \{1, 2, ..., N\}$. Clearly,

$$\mathbb{E}\sum_{q=1}^{N}\widehat{\phi}(v_i)_q^\top \widehat{\phi}(v_j)'_q = \frac{N}{r}\mathbb{E}\sum_{q\in\mathcal{S}_r}\widehat{\phi}(v_i)_q^\top \widehat{\phi}(v_j)'_q. \tag{24}$$

This shows that we can perform dimensionality reduction of GRFs by selecting 'anchor points' $\mathcal{S}_r$, whilst preserving unbiasedness. These arguments are unmodified when we also incorporate the extra $g$ variables. Note that, after randomly subsampling coordinates, the normalisation of each feature is increased by $\sqrt{N/r}$ to preserve unbiasedness.

**Labelled graphs.** Finally, we need to address graph node labels. Recall that, given a vertex labelling function $\mathcal{L} : \text{V}(\text{G}_1) \cup \text{V}(\text{G}_2) \to \text{L}$ (for a discrete set of labels L), the product graph is obtained by taking $\mathrm{V}(\mathrm{G}_1 \times \mathrm{G}_2) = \{(v_1, v_2) : v_1 \in \mathrm{V}(\mathrm{G}_1), v_2 \in \mathrm{V}(\mathrm{G}_2), \mathcal{L}(v_1) = \mathcal{L}(v_2)\}$, and $\mathrm{E}(\mathrm{G}_1, \mathrm{G}_2) = \{((v_1, v_2), (v_1', v_2')) : (v_1, v_1') \in \mathrm{E}(\mathrm{G}_1), (v_2, v_2') \in \mathrm{E}(\mathrm{G}_2), (v_1, v_2) \in \mathrm{V}(\mathrm{G}_1 \times \mathrm{G}_2), (v_1', v_2') \in \mathrm{V}(\mathrm{G}_1 \times \mathrm{G}_2)\}$. In other words, supervertices only exist in $\mathrm{G}_1 \times \mathrm{G}_2$ if their labels match. To account for this in our estimate of $\sum_{i=0}^{\infty} f_i \mathbf{A}^i_{\mathrm{G}_1 \times \mathrm{G}_2}$, we simply need to ablate contributions from our estimates of $\sum_{i_1=0}^{\infty} \sqrt{f_{i_1}} \mathbf{A}^{i_1}_{\mathrm{G}_1}$ and $\sum_{i_2=0}^{\infty} \sqrt{f_{i_2}} \mathbf{A}^{i_2}_{\mathrm{G}_2}$ where the sequence of node labels is not identical, since in this case the composite walk on $\mathrm{G}_1$ and $\mathrm{G}_2$ visits supervertices that do not exist and should not contribute to the sum. As discussed in Sec. 3.3 and shown in Alg. 1, this is straightforwardly achieved by multiplying by Rademacher $z$ variables that depend on each walker's present node label, $(z(n, l, w))_{n=1,\dots,N_L}^{w=0,\dots,m, l=0,1,\dots}$ with $N_L$ denoting the number of node labels. It is trivial to see that this generalises our algorithm to labelled graphs.

Having considered all cases, the proof is complete. □

### A.2 Proof of Thm. 4.2: GVoys give sharp kernel estimates

We now derive the concentration inequality for GVoys, presented in Eq. 12.

*Proof.* Initially, consider unlabelled graphs. Since all $m$ walks to share the same random number generator $(g(i))_{i=0}^{\infty}$, we have that

$$\mathbf{A}^{i_1}_{G_1} \otimes \mathbf{A}^{i_2}_{G_2} \mathbb{I}(i_1 = i_2) = \frac{1}{m^2} \sum_{w_1^{(i)}, w_2^{(i)}=1}^{m} \mathbb{E}\left( \widehat{\mathbf{A}}^{i_1}_{G_1}(w_1^{(i)}) \otimes \widehat{\mathbf{A}}^{i_2}_{G_2}(w_2^{(i)}) \right) \cdot \mathbb{E}\left( g(i_1) g(i_2) \right), \tag{25}$$

where $\widehat{\mathbf{A}}^{i_1}_{G_1}(w_1^{(i)})$ denotes the estimate of the $i_1$th power of adjacency matrix $\mathbf{A}_{G_1}$, constructed using GRFs by sampling random walk $w_1^{(i)}$. Assume that $r_i = N_i$, so every node is used as an anchor and there is no dimensionality reduction. Then note that

$$\mathbf{C}_1 = \frac{1}{m} \sum_{w_1^{(i)}=1}^{m} \sum_{i_1=0}^{\infty} \sqrt{f_{i_1}} \widehat{\mathbf{A}}^{i_1}_{G_1}(w_1^{(i)}) g(i_1). \tag{26}$$

Supposing that $\mathbf{C}_2$, $\mathbf{D}_1$ and $\mathbf{D}_2$ are computed analogously, we construct our RWK estimator

$$\widehat{\mathrm{K}}_{\mathrm{RWK}}(\mathrm{G}_1, \mathrm{G}_2) = (\mathbf{v}_1^\top \mathbf{C}_1 \mathbf{D}_1^\top \mathbf{w}_1)(\mathbf{v}_2^\top \mathbf{C}_2 \mathbf{D}_2^\top \mathbf{w}_2). \tag{27}$$

When Eq. 26 is constructed using Alg. 1 (but changing the normalisation $\frac{1}{\sqrt{m}} \to \frac{1}{m}$ because of the shared $g$ variables; see previous discussion), we take

$$[\mathbf{C}_1]_{q,:} = \frac{1}{m} \sum_{k=1}^{m} \zeta(\omega^{(q)}_{k,G_i}), \quad \text{with} \quad \zeta(\omega^{(q)}_{k,G_i}) \coloneqq \sum_{\omega \text{ p.s. } \omega_k^{(q)}} \sqrt{f_{\mathrm{len}(\omega)}} g(\mathrm{len}(\omega)) \left( \prod_{v \in \omega[:-1]} \frac{d_v}{\sqrt{1 - p_{\mathrm{halt}}}} \right) \boldsymbol{e}_{\omega[-1]}. \tag{28}$$

To remind the reader, here $\omega_k^{(q)}$ is the $k$th random walk (of a total of $m$) simulated out of node $q$ on graph $G_i$. $\omega$ p.s. $\omega_k^{(q)}$ means that $\omega$ is a *prefix subwalk* of $\omega_k^{(q)}$, meaning $\omega = \omega_k^{(q)}[: \mathrm{len}(\omega)]$. $\mathrm{len}(\omega)$ is the number of hops in the walk. $d_v$ denotes the degree of node $v$, and $p_{\mathrm{halt}}$ is the termination probability. $\omega[-1]$ is the last node of prefix subwalk $\omega$, and $\widehat{\boldsymbol{e}}_k$ is the the unit vector for coordinate $k$. $g(i)$ is the Rademacher random variable for step $i \in \{\mathbb{N} \cup 0\}$. Clearly, $\zeta(\omega^{(q)}_{k,G_i}) \in \mathbb{R}^N$.

Let us define the random variables

$$X_k \coloneqq \{\omega^{(q,\mathbf{C})}_{k,G_i}, \omega^{(q,\mathbf{D})}_{k,G_i}\}_{q \in \{1,\dots,N_i\}, G_i \in \{G_1, G_2\}} \quad \text{for} \quad k = 1, \dots, m. \tag{29}$$

Each variable contains a random walk out of every node on each of the two graphs to construct $\mathbf{C}$, and another independent copy for $\mathbf{D}$. Our strategy is to bound changes to the estimator when $X_k$ is modified.

Firstly, note that, from the properties of the random walk and Rademacher random variables,

$$\|\zeta(\omega^{(q)}_{k,G})\|_1 \leq \sum_{l=0}^{\infty} \sqrt{|f_l|} \left( \frac{\max_{v \in G_i} d_v}{\sqrt{1 - p_{\mathrm{halt}}}} \right)^k \eqqcolon c(\mathrm{G}). \tag{30}$$

Assume that $c(\mathrm{G}_1)$ and $c(\mathrm{G}_2)$ are finite, which is guaranteed for suitably regularised $f$. Observe that

$$[\mathbf{C}_1\mathbf{D}_1^\top]_{q_1q_2} = \frac{1}{m^2}\sum_{k_1=1}^{m}\sum_{k_2=1}^{m}\zeta(\omega_{k,G_1}^{(q_1)})^\top\zeta(\omega_{k,G_1}^{(q_2)}). \tag{31}$$

Supposing that *one of $X_k$ is modified*, we have that

$$|\Delta[\mathbf{C}_1\mathbf{D}_1^\top]_{q_1q_2}| \leq \frac{2(2m-1)}{m^2}c(\mathrm{G}_1)^2. \tag{32}$$

We can bound

$$|\mathbf{v}_1^\top\mathbf{C}_1\mathbf{D}_1^\top\mathbf{w}_1| \leq c(\mathrm{G}_1)^2\|\mathbf{v}_1\|_1\|\mathbf{w}_1\|_1 \tag{33}$$

and

$$|\Delta\left(\mathbf{v}_1^\top\mathbf{C}_1\mathbf{D}_1^\top\mathbf{w}_1\right)| \leq \frac{2(2m-1)}{m^2}c(\mathrm{G}_1)^2\|\mathbf{v}_1\|_1\|\mathbf{w}_1\|_1. \tag{34}$$

The same arguments hold for the equivalent quantities on $\mathrm{G}_2$. Lastly, we can bound

$$|\Delta\widehat{\mathrm{K}}(G_1,G_2)| \leq 2\left(\frac{4(2m-1)}{m^2}+\frac{4(2m-1)^2}{m^4}\right)\cdot c(\mathrm{G}_1)^2c(\mathrm{G}_2)^2\|\mathbf{v}_1\|_1\|\mathbf{w}_1\|_1\|\mathbf{v}_2\|_1\|\mathbf{w}_2\|_1 =: k(m,\mathrm{G}_1,\mathrm{G}_2). \tag{35}$$

We would like to apply McDiarmid's inequality using this bounded difference upon changing one of the $X_k$ variables. However, to obtain the result above we have assumed that the $g$ variables are shared across all of the $m$ walkers (in order to inherit a $\frac{1}{m}$ normalisation rather than $\frac{1}{\sqrt{m}}$ in Eq. 26), which means that they are not fully independent as $m$ grows. But the walkers are *conditionally* independent given a fixed draw of $g$. Hence, we can write

$$\Pr\left(|\widehat{\mathrm{K}}(G_1,G_2|g) - \mathbb{E}(\widehat{\mathrm{K}}(G_1,G_2|g))| \geq \epsilon\right) \leq 2\exp\left(-2\frac{\epsilon^2}{mk(m,\mathrm{G}_1,\mathrm{G}_2)^2}\right) \tag{36}$$

for any draw of $g$. Graph node labels are incorporated by multiplying loads by extra Rademacher variables, in our case also shared across all $m$ walkers. These extra $\pm 1$ factors will not modify our bounds on estimator differences, but will induce extra dependencies that need to be accounted for to ensure conditional independence. Therefore,

$$\Pr\left(|\widehat{\mathrm{K}}(G_1,G_2|g,z) - \mathbb{E}(\widehat{\mathrm{K}}(G_1,G_2|g,z))| \geq \epsilon\right) \leq 2\exp\left(-2\frac{\epsilon^2}{mk(m,\mathrm{G}_1,\mathrm{G}_2)^2}\right) \tag{37}$$

for any $g, z$, as claimed. □

### A.3 Proof of Thm. 4.3: Rademacher is optimal

In this section, we prove that Rademacher random variables are the optimal choice for the random variables $g(l,w)$. Recall that for unbiasedness we require that $\mathbb{E}(g(l,w)) = 0$ and $\mathbb{E}(g(l,w)^2) = 1$.

*Proof.* Initially, suppose that the graphs are unlabelled. The estimator for the RWK can be written

$$\widehat{\mathrm{K}}_{\mathrm{RWK}}(\mathrm{G}_1,\mathrm{G}_2) = \frac{1}{m^2}\sum_{w_1^{(i)},w_2^{(i)},w_1^{(j)},w_2^{(j)}=1}^{m}\sum_{i_1,i_2,j_1,j_2=0}^{\infty}\widehat{\mathrm{M}}_{G_1}(i_1,j_1,w_1^{(i)},w_1^{(j)})\widehat{\mathrm{M}}_{G_1}(i_2,j_2,w_2^{(i)},w_2^{(j)}) \cdot g(i_1,w_1^{(i)})g(i_2,w_2^{(i)})g'(j_1,w_1^{(j)})g'(j_2,w_2^{(j)}) \tag{38}$$

where we defined

$$\widehat{\mathrm{M}}_{G_1}(i_1,j_1,w_1^{(i)},w_1^{(j)}) \coloneqq \mathbf{v}_1^\top\left[\left(\sqrt{f_{i_1}}\widehat{\mathbf{A}}_{\mathrm{G}_1}^{i_1}(w_1^{(i)})\right)\left(\sqrt{f_{j_1}}\widehat{\mathbf{A}}_{\mathrm{G}_1}'^{j_1}(w_1^{(j)})\right)\right]\mathbf{w}_1, \tag{39}$$

$$\widehat{\mathrm{M}}_{G_2}(i_2,j_2,w_2^{(i)},w_2^{(j)}) \coloneqq \mathbf{v}_1^\top\left[\left(\sqrt{f_{i_2}}\widehat{\mathbf{A}}_{\mathrm{G}_2}^{i_2}(w_2^{(i)})\right)\left(\sqrt{f_{j_2}}\widehat{\mathbf{A}}_{\mathrm{G}_2}'^{j_2}(w_2^{(j)})\right)\right]\mathbf{w}_2. \tag{40}$$

Here, $\widehat{\mathbf{A}}_{\mathrm{G}_1}^{i_1}(w_1^{(i)})$ denotes the estimate of the $i_1$th power of $\mathbf{A}_{\mathrm{G}_1}$, obtained using GRFs by simulating random walk $w_1^{(i)}$. This expression is the same as 27, but pulls out the sequences $(g(i,w))_{w=1,\dots,m}^{i=0,\dots,\infty}$ and $(g'(i,w))_{w=1,\dots,m}^{i=0,\dots,\infty}$ explicitly for clarity.

The variance depends on the expectation of the squared estimator, which introduces extra summations. We have that:

$$\mathbb{E}\left(\widehat{\mathrm{K}}_{\mathrm{RWK}}(\mathrm{G}_1,\mathrm{G}_2)^2\right) = \frac{1}{m^4} \sum_{\substack{w_1^{(i)},w_2^{(i)},w_1^{(j)},w_2^{(j)},\\ w_1'^{(i)},w_2'^{(i)},w_1'^{(j)},w_2'^{(j)}=1}}^{m} \sum_{\substack{i_1,i_2,j_1,j_2\\ i_1',i_2',j_1',j_2'=0}}^{\infty} \mathbb{E}\Big[\widehat{\mathrm{M}}_{G_1}(i_1,j_1,w_1^{(i)},w_1^{(j)})\widehat{\mathrm{M}}_{G_2}(i_2,j_2,w_2^{(i)},w_2^{(j)}) \\ \widehat{\mathrm{M}}_{G_1}(i_1',j_1',w_1'^{(i)},w_1'^{(j)})\widehat{\mathrm{M}}_{G_2}(i_2',j_2',w_2'^{(i)},w_2'^{(j)})\Big] \cdot \\ \mathbb{E}\left(g(i_1,w_1^{(i)})g(i_2,w_2^{(i)})g(i_1',w_1'^{(i)})g(i_2',w_2'^{(i)})\right)\mathbb{E}\left(g'(j_1,w_1^{(j)})g'(j_2,w_2^{(j)})g'(j_1',w_1'^{(j)})g'(j_2',w_2'^{(j)})\right). \tag{41}$$

Assume that $\mathbb{E}\left[\widehat{\mathrm{M}}_{G_1}(i_1,j_1,w_1^{(i)},w_1^{(j)})\widehat{\mathrm{M}}_{G_2}(i_2,j_2,w_2^{(i)},w_2^{(j)})\widehat{\mathrm{M}}_{G_1}(i_1',j_1',w_1'^{(i)},w_1'^{(j)})\widehat{\mathrm{M}}_{G_2}(i_2',j_2',w_2'^{(i)},w_2'^{(j)})\right] \geq 0$ for all arguments, which holds e.g. if all the entries of $\{\mathbf{v}_1,\mathbf{v}_2,\mathbf{w}_1,\mathbf{w}_2\}$ are greater than or equal to 0 since all estimators of powers of the adjacency matrices are positive. The typical choice is $\mathbf{w}=\mathbf{v}=\mathbf{1}_{N_1N_2}$, so this condition is reasonable. Given the zero mean and unit variance properties, the only nonzero terms in the expectation that depend on $g$ are proportional to $\mathbb{E}\left(g(i,w)^4\right)$ or $\mathbb{E}\left(g(i,w)^4\right)^2$. Note that $\mathbb{E}\left(g(i_1,w)^4\right) \geq \mathbb{E}\left(g(i_1,w)^2\right)^2 = 1$. For the Rademacher random variable $g(i_1,w)\in\{-1,+1\}$ with equal probability, we have that $\mathbb{E}\left(g(i_1,w)^4\right)=1$, so this is exactly minimised. Then the product is also minimised, so every term in $\mathbb{E}\left(\widehat{\mathrm{K}}_{\mathrm{RWK}}(\mathrm{G}_1,\mathrm{G}_2)^2\right)$ is minimised. Thus, the variance is minimised.

Now suppose that the graph is labelled, so we also incorporate extra $z$ Rademacher random variables at every timestep that depend on the particular graph node. The expectation of products of these variables will only ever evaluate to 1 or 0, so when we incorporate them into Eq. 41 the only terms that survive will still be proportional to $\mathbb{E}\left(g(i,w)^4\right)$ or $\mathbb{E}\left(g(i,w)^4\right)^2$. Hence, the same arguments still hold; Rademacher is optimal for $g$. This completes the proof. □

# B ADDITIONAL EXPERIMENTAL DETAILS

In this section we provide additional details for our experiments. First we present the statistics for various datasets used for our graph classification task. All our experiments are run on a Ryzen 7 with 32 GB memory.

## B.1 Convergence Experiments

For these experiments, we use a halting probability of .2, and $\mu_i = \frac{\lambda^i}{i!}$. $\mathbf{v},\mathbf{w}$ are chosen to be uniform probability distribution on the nodes of the product graph.To ensure convergence, $\lambda$ is chosen as $\frac{1}{\mathrm{d}_{\max}^2}$, where $\mathrm{d}_{\max}$ is the maximum degree among all vertices for the pair of graphs. We run our experiments by choosing a pair of graphs from NC1, ENZYMES, PROTEINS and DD. The size of the graphs ranged from 20 to 180 nodes and edges from 50 to almost 900. Each experiment was run 10 times with different random seeds.

## B.2 Speed Experiments

We first detail the experimental setup for the speed experiments. We create a dataset of 10 Erdos-Renyi graphs with sizes $2^k$, where $k=1,\cdots 14$, and the probability of edges between the nodes is .1. Moreover, we made sure that the graphs are connected. $\mathbf{v},\mathbf{w}$ are chosen to be uniform probability distribution on the nodes of the product graph. To ensure convergence, $\lambda$ is chosen as $\frac{1}{\mathrm{d}_{\max}^2}$, where $\mathrm{d}_{\max}$ is the maximum degree among all vertices for all the graphs in the dataset. The number of walks in GVoys is set to 100 and the blocks to be 10 for the block-GVoy variant. For both the variants, the halting probability is set to be .2. For conjugate gradient and the fixed point methods, we set the maximum number of iterations to be 150k with a tolerance threshold to be $10^{-6}$.

## B.3 Graph Classification

First we present the statistics for the datasets used for our graph classification task. For all the experiments, we use the number of random walks to be 1000, halting probability to be .2. $\mathbf{v},\mathbf{w}$ are chosen to be uniform probability distribution on the nodes of the product graph. We follow the same setup using a kernel-SVM and the same $\lambda$ as in Nikolentzos et al. (2021) and report the 10-fold cross validation results.

Table 2: Statistics of Graph Classification Datasets used in this paper

| Datasets | # Graphs | # Labels | Avg. # Nodes | Avg. # Edges | # Node Labels | # Node Attributes |
|---|---|---|---|---|---|---|
| Mutag | 188 | 2 | 17.93 | 19.79 | 7 | - |
| Ptc-Mr | 344 | 2 | 14.29 | 14.69 | 19 | - |
| Enzymes | 600 | 6 | 32.63 | 62.14 | 3 | 18 |
| Proteins | 1113 | 2 | 39.06 | 72.82 | 3 | 1 |
| D&D | 1178 | 2 | 284.32 | 715.66 | 82 | - |
| NCI1 | 4110 | 2 | 29.87 | 32.30 | 37 | - |
| AIDS | 2000 | 2 | 15.69 | 16.20 | 38 | 4 |

### B.4 Learning RWK

For this experiment, we choose $\mathbf{w}, \mathbf{v}$ (in Eq. 1) to be $\mathbf{1}$. Our goal in this experiment is to learn a set of $\mu_i$, which can beat the baseline kernel in downstream tasks. We truncate Eq 1 to 8 terms. We learn the $\mu_i$ (which we constrain to be positive) along with the weights of a SVM in an end-to-end manner. We observe that a kernel which is learned in this manner outperforms the baseline kernel and the weights do not show any particular pattern (for instance decreasing), showing the importance to be able to customize the $\mu$'s for various downstream applications.

## C ADDITIONAL RESULTS

In this section, we present additional results for our GVoys algorithm.

### C.1 Convergence Results

In this subsection, we present 2 additional results showing that GVoys converge to the true kernel as the number of random walks gets large. Fig 5 shows our results on mutag and on ptc-mr datasets. The hyperparamters for this experiment is same as those detailed in Appendix B.1.

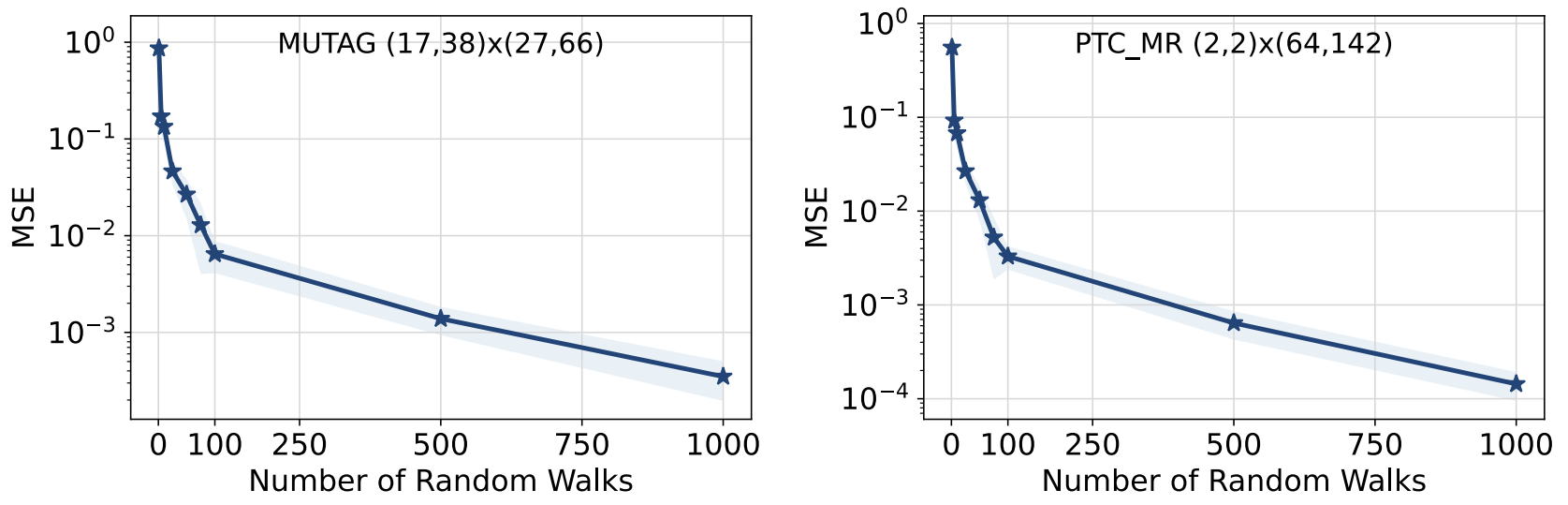

Figure 5: Comparing the approximation error of GVoys on samples of pairs of graphs from various datasets from TUDataset (Morris et al., 2020) as a function of number of random walks. The pair of tuples next to each dataset name corresponds to the number of vertices and edges of the respective graphs. Shaded regions represent std-devs (over 10 runs).

### C.2 Speed as a function of number of random walks

In this section, we compare the speed of GVoys as a function of the number of random walks with the *exact* RWK (using Equation 1). Fig 6 presents the time as a function of walks.

The graphs for this experiment are taken from Protein and DD datasets from the benchmark TUDatasets (Morris et al., 2020). The pair of tuples next to each dataset name corresponds to the number of nodes and edges of the respective graphs. For this task, a pair of graphs from each of the datasets with nodes ranging from $100 - 180$ and edges ranging from $500 - 850$. For both pairs of graphs, our method is almost $\mathbf{10}\times$ faster than the baseline.

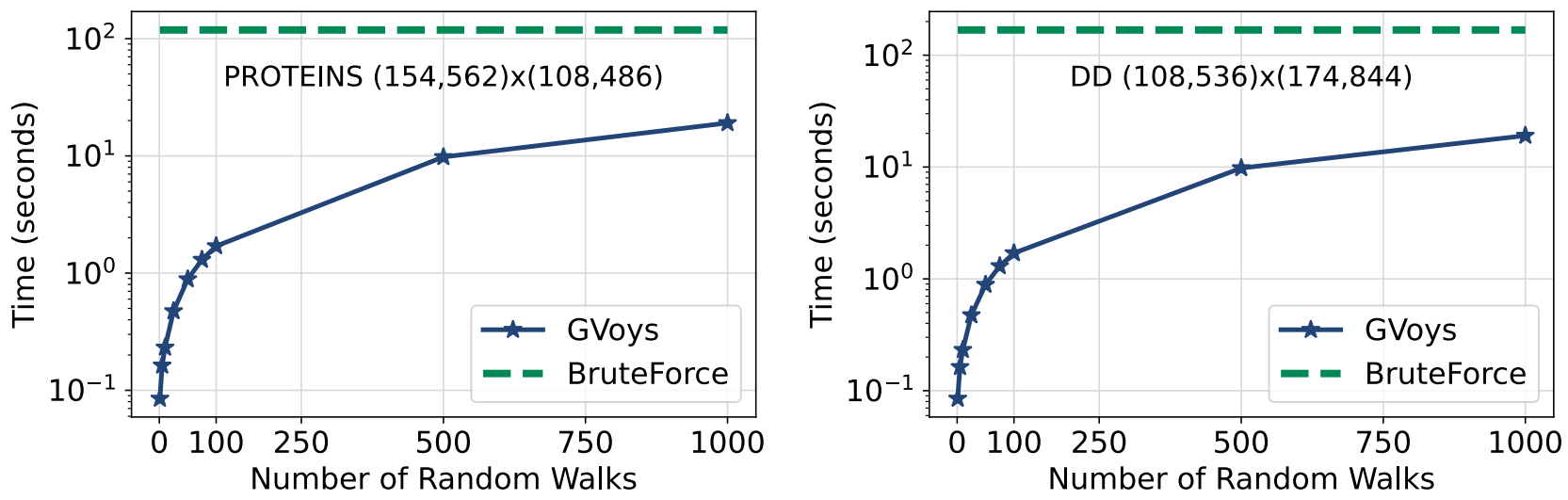

Figure 6: Runtime of GVoys as a function of the number of random walks on samples of pairs of graphs from PROTEIN and DD datasets. GVoys are considerably faster than the RWK even with 1000 random walks. The pair of tuples next to each dataset name corresponds to the number of nodes and edges of the respective graphs.

### C.3 Learning RWK Kernels

In this section, we show an additional experiment where learning $\mu$ can provide gains over the baseline geometric RWK kernel. We use the ENZYMES dataset from TUDataset (Morris et al., 2020) and report the 10-fold accuracy. In this case, we learn 8 $\mu$ coefficients in Eq 1 in an end-to-end manner. We fit this trainable kernel in a single layer neural network (simply a linear layer) which is trained using cross-entropy loss for 200 iterations on each fold. We obtain accuracy scores of $20.7 \pm 4.2$, whereas the baseline geometric RWK trained using the same cross-entropy loss gives $18.1 \pm 3.4$, a relative improvement of **14%**. Moreover, we do not notice any decreasing behavior among the learned 8 coefficients. For our experiment, the $\lambda$ is learned along with $\mu$, and for the baseline $\lambda$ is the $1/d_{\max}^2$, where $d_{\max}$ is the maximum degree among all graphs in the dataset. This experiment illustrates the need for flexibility for general RWK.

### C.4 Point Cloud Transformer (PCT) Experiments

We have run additional experiments to test GVOYs. We augmented regular Point Cloud Transformer (four attention layers, one head per layer, query/key dimensionality = 128) with the embedding of the corresponding graph, obtained via GVOY-algorithm. The graph was created by first computing dot-product similarity matrix, leveraging (x,y,z)-positions of the points and then thresholding (for large enough dot-product we put an edge between the corresponding points in the constructed graph). The embedding was then added to the original PCT embedding. On the semantic segmentation task for the Stanford Large-Scale 3D Indoor Spaces Dataset, we obtained 3% accuracy improvement.